\documentclass{article}

\PassOptionsToPackage{numbers}{natbib}
 \usepackage[main, final]{neurips_2026}

\usepackage[utf8]{inputenc} 
\usepackage[T1]{fontenc}    
\usepackage{hyperref}       
\usepackage{url}            
\usepackage{booktabs}       
\usepackage{amsfonts}       
\usepackage{nicefrac}       
\usepackage{microtype}      
\usepackage{xcolor}         

\usepackage{enumitem}
\usepackage{graphicx}
\usepackage{amsmath}
\usepackage{wrapfig}
\usepackage{array}
\usepackage{tabularx}
\usepackage{makecell}
\usepackage{float}
\usepackage{subcaption} 
\title{WorldComposer: Generative High-Fidelity Simulation with Digital Cousins for Generalizable Robot Learning and Evaluation}

\author{
    \textbf{Jasper Lu}\textsuperscript{1*},
    \textbf{Zhenhao Shen}\textsuperscript{1*},
    \textbf{Yuanfei Wang}\textsuperscript{1*},
    \textbf{Shugao Liu}\textsuperscript{1},
    \textbf{Shengqiang Xu}\textsuperscript{1},
    \textbf{Shawn Xie}\textsuperscript{1}, \\
    \textbf{Jingkai Xu}\textsuperscript{1},
    \textbf{Feng Jiang}\textsuperscript{1},
    \textbf{Jade Yang}\textsuperscript{1},
    \textbf{Chen Xie}\textsuperscript{3},
    \textbf{Ruihai Wu}\textsuperscript{2,1$\dagger$} \\[0.5em] 
    \textsuperscript{1}Peking University \quad \textsuperscript{2}University of California, Berkeley \quad \textsuperscript{3}Lightwheel \\
    *Equal contribution, $\dagger$Corresponding author \quad Contact email: {\tt ruihai@berkeley.edu} \\[0.5em]
    Webpage: \href{https://stubborn111.github.io/WorldComposer/}{https://stubborn111.github.io/WorldComposer/}
    \vspace{-1em} 
}

\begin{document}

\maketitle

\begin{abstract}
Learning robust robot policies in real-world environments requires diverse data augmentation, yet scaling real-world data collection is costly due to the need for acquiring physical assets and reconfiguring environments. Therefore, augmenting real-world scenes into simulation has become a practical augmentation for efficient learning and evaluation. We present \textbf{WorldComposer}, an automated generative Real2Sim2Real framework that maps real-world panoramas into high-fidelity, task-ready simulation environments and further synthesizes diverse Digital Cousins through scene and object variations. Combined with high-quality physics engines and realistic assets, WorldComposer supports interactive manipulation across rigid, articulated, and deformable objects. Additionally, we incorporate multi-room stitching to construct consistent large-scale environments for long-horizon tasks. Experiments demonstrate a strong sim-to-real correlation validating our platform's fidelity, and show that extensively scaling up data generation leads to significantly better generalization to unseen scene and object variations, demonstrating the effectiveness of Digital Cousins for generalizable robot learning and evaluation.
\end{abstract}

\section{Introduction}
\label{sec:intro}

Robust robot learning in real-world environments depends heavily on access to diverse and realistic training data~\cite{intelligence2025pi05visionlanguageactionmodelopenworld, bjorck2025gr00t, cheang2025gr3technicalreport}. In particular, manipulation policies are highly sensitive to variations in scene layout, object geometry, appearance, and physical interactions. Collecting such diverse data directly in the real world, however, remains costly and labor-intensive, as it typically requires acquiring additional physical assets, repeatedly reconfiguring environments, and conducting extensive trials under varied conditions. These challenges significantly affect the policy robustness and generalization.

Simulation provides a practical augmentation by enabling efficient data generation and controlled experimentation. With modern physics engines~\cite{NVIDIA_Isaac_Sim, genesis2026genesisworld} and increasingly realistic assets~\cite{li2025lehome,chen2025robotwin}, simulation has become a valuable tool for robot learning. Nevertheless, constructing simulation environments that faithfully reflect real-world scenes remains challenging. Existing approaches often rely on manual scene modeling, predefined templates, or limited domain randomization. While effective in constrained settings, these methods either require substantial human effort or fail to capture the structural and semantic complexity of real-world environments.


Recent advances ~\cite{marble2025,hyworld22026, gigaworldteam2025gigaworld0worldmodelsdata, jiang2025gsworldclosedloopphotorealisticsimulation} in generative models offer new opportunities to bridge the gap between real-world perception and simulation. By leveraging visual observations, it becomes possible to automatically reconstruct simulation environments from real-world data. However, most prior work focuses on generating a single digital replica of a scene, providing limited support for systematic variation and data augmentation. As a result, the generated environments often lack the diversity needed to train and evaluate policies that generalize across unseen scene and object variations.

\begin{figure}[t]
\centering
\includegraphics[width=\textwidth]{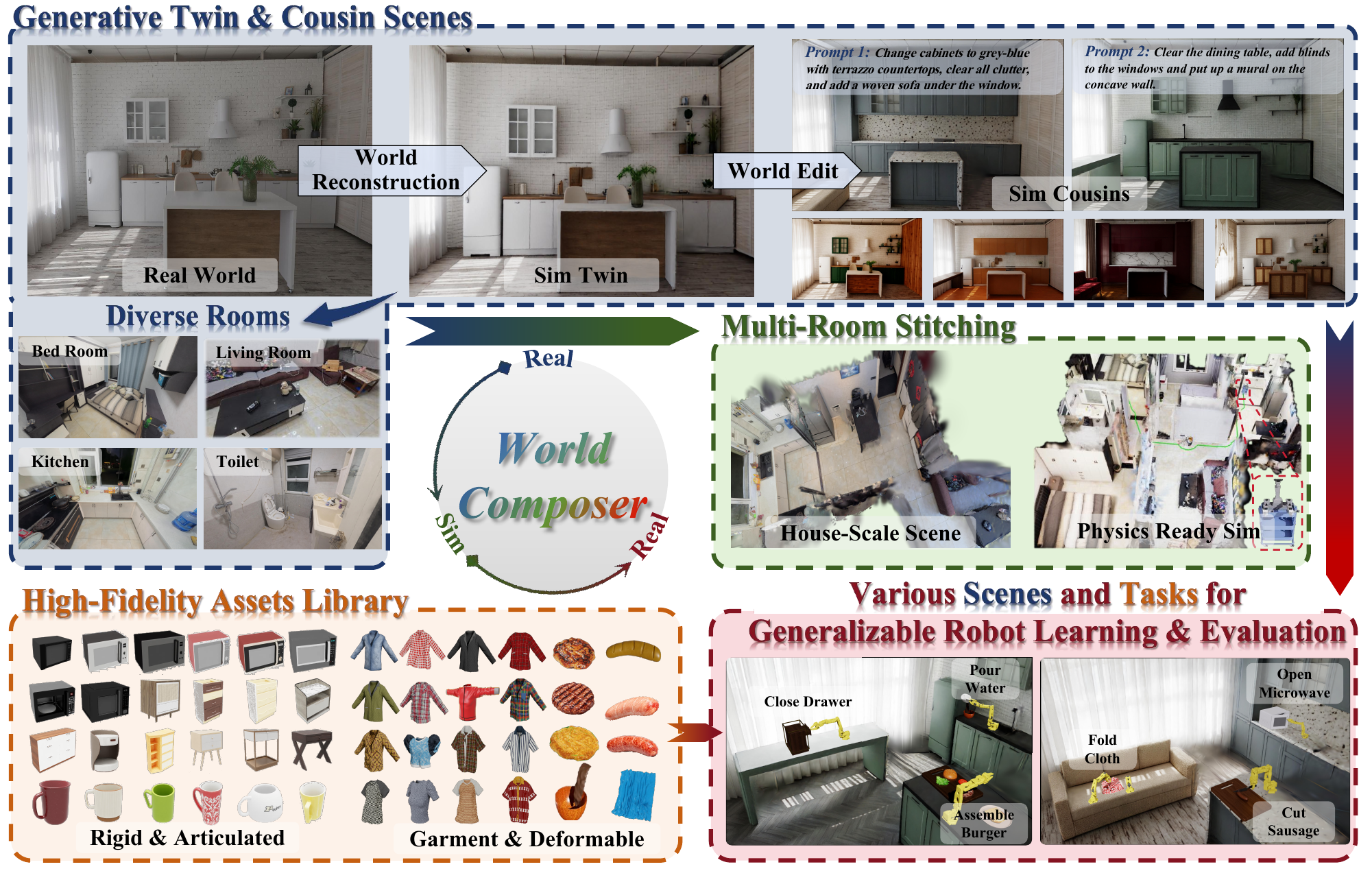}
\caption{\textbf{WorldComposer} generates high-fidelity simulation scenes from real-world panorama, and further generates diverse cousin scenes through editing. The generated rooms can be seamlessly stitched together into multi-room environments for navigation. Combined with interactive tasks, we provide a platform for generalizable policy learning and evaluation.}
\label{fig:teaser}
\vspace{-2mm}
\end{figure}


In this paper, as illustrated in Fig.~\ref{fig:teaser}, we present \textbf{WorldComposer}, an automated generative Real2Sim2Real framework that transforms real-world captures into high-fidelity, task-ready simulation environments. By combining generative scene reconstruction with real-to-sim alignment and interactive simulation, WorldComposer provides an efficient pathway from real-world observations to robot-ready simulation. The framework supports multi-room scene construction, enabling complex environments suitable for scene-level tasks like navigation. Crucially, our framework operates with high-quality physics engines~\cite{NVIDIA_Isaac_Sim, li2025lehome} and realistic asset libraries spanning diverse object categories and interaction types, enabling interactive manipulation tasks beyond static scene generation.

Moreover, our framework supports efficient generation of diverse digital cousins of both scenes and objects through semantic and geometric editing. These digital cousins introduce controlled variations in layout, geometry, and object configurations, significantly expanding the diversity of interactive experiences available for training. By augmenting both environments and assets, our approach provides rich data for improving policy robustness and generalization.


The generated environments form a platform for generalizable robot learning and evaluation. By training and evaluating policies across diverse digital cousins, we enable systematic assessment of policy generalization under controlled scene and object variations. Extensive experiments demonstrate that policies trained on our platform generalize better to unseen scenes and objects. Moreover, we observe strong correlation between simulation and real-world performance, indicating our platform provides reliable basis for both learning and evaluation. In summary, our contributions are as follows:
\begin{itemize}[leftmargin=*]
    \item \textbf{Automated Generative Real2Sim2Real Framework.}
    We present \textbf{WorldComposer}, an automated framework that converts real-world panoramas into high-fidelity, task-ready simulation environments to support real-robot policy learning and evaluation.
    \item \textbf{Generalizable Robot Learning and Evaluation Platform.}
    WorldComposer expands Digital Twins into scalable Digital Cousins through scene and object variations, and automatically collects task trajectories with parameterized scripted skills. These components turn finite real-world captures into diverse simulation data for policy learning and controlled evaluation.
    \item \textbf{Comprehensive Experimental Validation.}
    Extensive experiments show that policies trained on our platform achieve improved generalization across unseen scenes and objects. Besides, results in simulation exhibit strong correlation with real-world performance.
    
\end{itemize}





\begin{table}[t]
\caption{Structural comparison against related simulation frameworks.}
\label{tab:framework_comparison}
\centering
\renewcommand{\arraystretch}{1.1}

\resizebox{\textwidth}{!}{
\begin{tabular}{l c c c c c c}
\toprule
\textbf{}%
{\textbf{Method}}
& \textbf{Real2Sim2Real}
& \begin{tabular}[c]{@{}c@{}}
    \textbf{Visual Fidelity}\\
    \textbf{(GS)}
  \end{tabular}
& \begin{tabular}[c]{@{}c@{}}
    \textbf{Physics Fidelity}\\
    \textbf{Support}
  \end{tabular}
& \begin{tabular}[c]{@{}c@{}}
    \textbf{Digital}\\
    \textbf{Cousins}
  \end{tabular}
& \begin{tabular}[c]{@{}c@{}}
    \textbf{Auto Scene}\\
    \textbf{Gen.}
  \end{tabular}
& \begin{tabular}[c]{@{}c@{}}
    \textbf{Auto Data}\\
    \textbf{Collection}
  \end{tabular} \\
\midrule

\textbf{RoboGen}~\cite{wang2023robogen}
& \textcolor{red!80!black}{$\times$}
& \textcolor{red!80!black}{$\times$}
& Rigid \& Articulated
& \textcolor{red!80!black}{$\times$}
& \textcolor{green!60!black}{$\checkmark$}
& \textcolor{green!60!black}{$\checkmark$} \\

\textbf{SaR Co-Training}~\cite{Maddukuri2025SimandRealCA}
& \textcolor{red!80!black}{$\times$}
& \textcolor{red!80!black}{$\times$}
& Rigid \& Articulated
& \textcolor{green!60!black}{$\checkmark$}
& \textcolor{red!80!black}{$\times$}
& \textcolor{red!80!black}{$\times$} \\

\textbf{URDFormer}~\cite{chen2024urdformerpipelineconstructingarticulated}
& \textbf{\textcolor{green!60!black}{$\boldsymbol{\checkmark}$}}
& \textcolor{red!80!black}{$\times$}
& Rigid \& Articulated
& \textcolor{red!80!black}{$\times$}
& \textcolor{green!60!black}{$\checkmark$}
& \textcolor{red!80!black}{$\times$} \\

\textbf{GSWorld}~\cite{jiang2025gsworldclosedloopphotorealisticsimulation}
& \textbf{\textcolor{green!60!black}{$\boldsymbol{\checkmark}$}}
& \textbf{\textcolor{green!60!black}{$\boldsymbol{\checkmark}$}}
& Rigid
& \textcolor{red!80!black}{$\times$}
& \textcolor{green!60!black}{$\checkmark$}
& \textcolor{red!80!black}{$\times$} \\

\textbf{SplatSim}~\cite{qureshi2024splatsimzeroshotsim2realtransfer}
& \textbf{\textcolor{green!60!black}{$\boldsymbol{\checkmark}$}}
& \textbf{\textcolor{green!60!black}{$\boldsymbol{\checkmark}$}}
& Rigid
& \textcolor{red!80!black}{$\times$}
& \textcolor{green!60!black}{$\checkmark$}
& \textcolor{green!60!black}{$\checkmark$} \\

\textbf{DISCOVERSE}~\cite{DISCOVERSE}
& \textbf{\textcolor{green!60!black}{$\boldsymbol{\checkmark}$}}
& \textbf{\textcolor{green!60!black}{$\boldsymbol{\checkmark}$}}
& Rigid \& Articulated
& \textcolor{red!80!black}{$\times$}
& \textcolor{green!60!black}{$\checkmark$}
& \textcolor{red!80!black}{$\times$} \\

\textbf{GS-Playground}~\cite{jia2026gsplaygroundhighthroughputphotorealisticsimulator}
& \textbf{\textcolor{green!60!black}{$\boldsymbol{\checkmark}$}}
& \textbf{\textcolor{green!60!black}{$\boldsymbol{\checkmark}$}}
& Rigid
& \textcolor{red!80!black}{$\times$}
& \textcolor{green!60!black}{$\checkmark$}
& \textcolor{red!80!black}{$\times$} \\

\midrule
{\textbf{WorldComposer}}
& \textbf{\textcolor{green!60!black}{$\boldsymbol{\checkmark}$}}
& \textbf{\textcolor{green!60!black}{$\boldsymbol{\checkmark}$}}
& \begin{tabular}[c]{@{}c@{}}
    Rigid, Articulated,\\
    Deformable, Fluid
  \end{tabular}
& \textbf{\textcolor{green!60!black}{$\boldsymbol{\checkmark}$}}
& \textbf{\textcolor{green!60!black}{$\boldsymbol{\checkmark}$}}
& \textbf{\textcolor{green!60!black}{$\boldsymbol{\checkmark}$}} \\
\bottomrule
\end{tabular}
}
\end{table}

\section{Related Work}

\subsection{Multimodal World Models}
\label{subsec:related_worldmodels}

Recent advancements in world model have been propelled by large-scale video generation models~\cite{kong2025hunyuanvideosystematicframeworklarge,openai2024sora,openai2025sora2,wan2025wanopenadvancedlargescale,nvidia2025worldsimulationvideofoundation, chen2025largevideoplannerenables}. However, these models lack explicit 3D geometry and physics required for precise robotic planning. Therefore, emerging works attempt to lift 2D generation into 3D space~\cite{kim2025videofrom3d3dscenevideo,ren2025gen3c3dinformedworldconsistentvideo, gigaworldteam2025gigaworld0worldmodelsdata, zhen2025tesseractlearning4dembodied}. Furthermore, generative 3D methods\cite{marble2025,hyworld22026} directly synthesize consistent, editable environments. Leveraging these models for scalable scene generation, we bridge the interactivity gap by populating environments with physics-enabled 
assets to construct interactive ``Digital Cousins.''

\subsection{High-Fidelity Simulation Platforms}
\label{subsec:related_sim}

To facilitate sim-to-real transfer, simulation environments have advanced in \textbf{visual} and \textbf{physical} fidelity. For visual realism, platforms have shifted from procedural environments~\cite{james2020rlbench,liu2023libero,szot2021habitat, li2021igibson} to photo-realistic scenes using digital twins~\cite{Mu_2025_CVPR, chen2025robotwin} and advanced rendering~\cite{nasiriany2024robocasa,li2023behavior,geng2025roboverse, guo2025genesismultimodaldrivingscene}. Simultaneously, physical fidelity has evolved beyond rigid-body dynamics to model fluids~\cite{xian2023fluidlab}, thin-shells~\cite{wang2023thin, seita2023learningrearrangedeformablecables, lu2024garmentlab} and volumetric objects~\cite{huang2021plasticinelab,heiden2021disect}. Our pipeline combines 3DGS-based reconstruction with rigorous hybrid physics to generate interactive and high-fidelity environments.

\subsection{Real-to-Sim Environment Creation}
\label{subsec:related_real2sim}

The field of Real-to-Sim has evolved from procedural layout generation~\cite{deitke2022phone2procbringingrobustrobots} to photorealistic digitization powered by 3D Gaussian Splatting (3DGS)~\cite{kerbl20233dgaussiansplattingrealtime}. 
To enable rich contact interactions, recent studies have integrated 3DGS with physics engines, creating closed-loop simulators that function as effective ``Digital Twins'' for robotic manipulation~\cite{qureshi2024splatsimzeroshotsim2realtransfer, jiang2025gsworldclosedloopphotorealisticsimulation, li2025robogsimreal2sim2realroboticgaussian, han2025re3simgeneratinghighfidelitysimulation, jia2026gsplaygroundhighthroughputphotorealisticsimulator}. 
Concurrently, these high-fidelity environments are increasingly leveraged as scalable benchmarks for policy evaluation, demonstrating strong correlations between simulation  and real-world performance~\cite{jain2025polarisscalablerealtosimevaluations, abouchakra2025realissimbridgingsimtorealgap, dan2025xsimcrossembodimentlearningrealtosimtoreal, pfaff2025scalablereal2simphysicsawareasset}. 
Our work extends this trajectory by introducing a generative pipeline that expands static twins into diverse ``Digital Cousins,'' serving as both a scalable engine for data augmentation and a rigorous testbed for generalization evaluation. Table~\ref{tab:framework_comparison} provides a systematic comparison of these frameworks.

\begin{figure*}[th]
\centering
\includegraphics[width=1.0\textwidth]{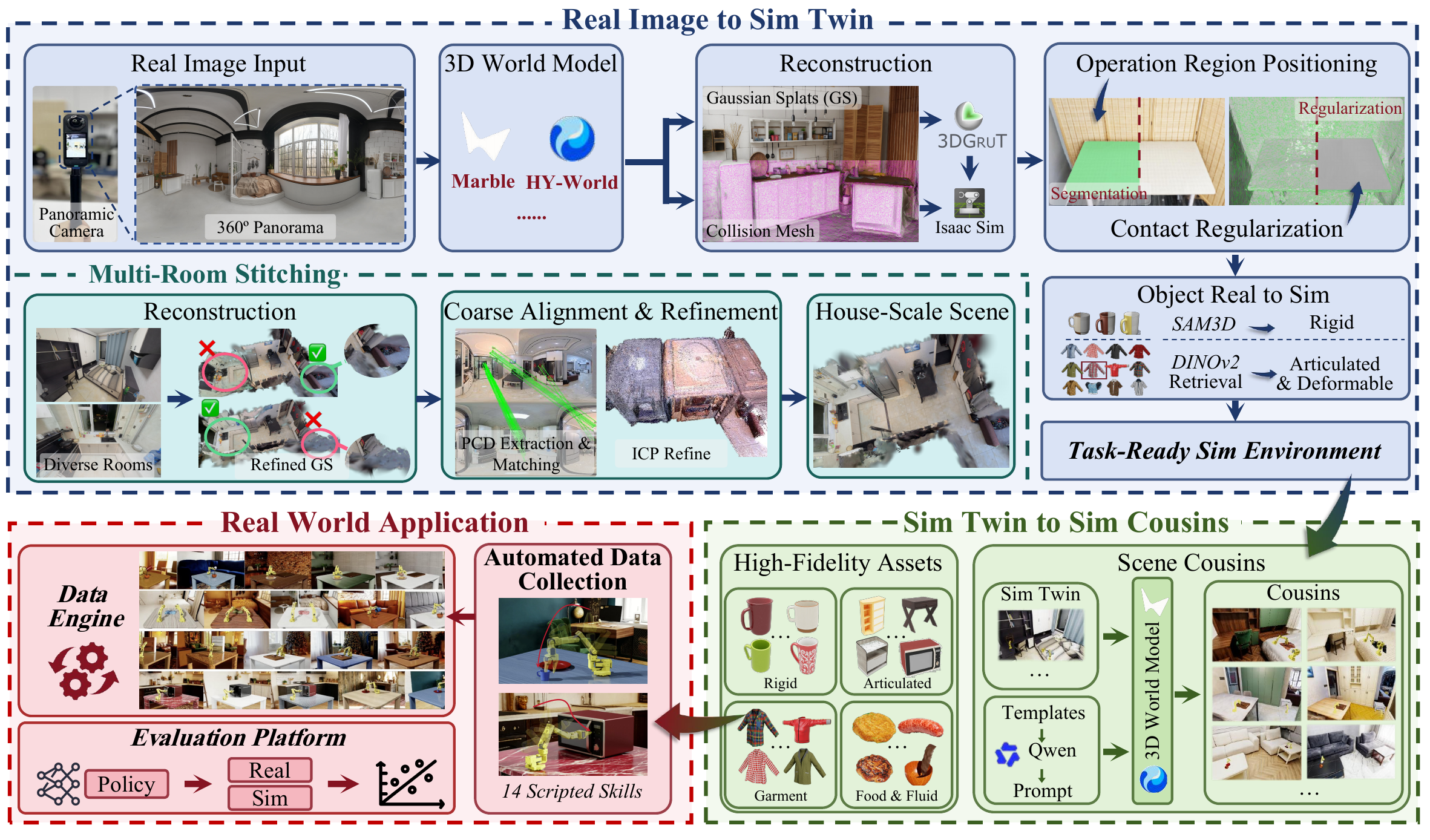} 
\caption{
\textbf{Overview of WorldComposer.}
WorldComposer turns real-world panoramic observations into task-ready Sim Twins, expands them into diverse Sim Cousins, and serves as both a trajectory-generation engine for robot learning and a high-fidelity platform for controlled evaluation.  
}
\label{fig:Pipeline}
\vspace{-0.5cm}
\end{figure*}

\section{Method}
\label{sec:method}

As illustrated in Fig.~\ref{fig:Pipeline}, \textbf{WorldComposer} is an automated generative Real2Sim2Real framework for generalizable robot learning and evaluation.
First, WorldComposer constructs task-ready Digital Twins by reconstructing scenes and matching real-world objects to high-fidelity physics-enabled assets (Section~\ref{subsec:scene_gen}), and stitches multiple rooms into house-scale navigable environments (Section~\ref{subsec:multi_room}).
    It then expands Digital Twins into scalable Digital Cousins and automatically collects task trajectories (Section~\ref{subsec:GRLE}), providing both a data engine for policy learning and a high-fidelity evaluation platform.

\subsection{Automated Real-to-Sim Task Construction}
\label{subsec:scene_gen}

The first stage of WorldComposer transforms real-world panorama into task-ready Digital Twin through a coarse-to-fine construction process, from visual and geometric scene reconstruction to real-to-sim alignment, contact regularization, and automatic task construction. 

\textbf{From Panorama to Digital Twin}.
Given a real-world 360-degree panorama, WorldComposer uses a generative and reconstruction-based 3D world model to reconstruct scene geometry and appearance. In our implementation, this module is instantiated with Marble~\cite{marble2025}, and can be extended to other 3D world models such as HY-World 2.0~\cite{hyworld22026} when compatible visual and geometric outputs are available. The reconstruction produces two key components: (i) \textit{Visual Rendering}, a set of 3D Gaussian Splats that enables photorealistic rendering while capturing complex lighting and material effects; and (ii) \textit{Collision Geometry}, a corresponding mesh that supports robot-environment interaction by providing rigid bodies for physics simulation. To integrate these components into the Isaac Sim simulator, we utilize \textit{3DGRUT}~\cite{wu20253dgut} to convert the Gaussian PLY data into the USD format, and import the collision mesh as physics-enabled geometry, ensuring compatibility with the Omniverse ecosystem.

\textbf{Operation Region Positioning and Contact Regularization.}
The generated scene remains a raw reconstruction, lacking explicit operation-region semantics and containing rough contact surfaces. To make it task-ready, we render views from the generated scene, use SAM3~\cite{carion2025sam3segmentconcepts} to segment candidate support regions, and apply RANSAC~\cite{Ransac} to fit the largest planar workspace. The corresponding collision mesh is regularized toward the fitted plane, reducing local height variation (\(\leq 1\) cm) and producing a stable support surface for downstream manipulation task construction.


\textbf{Automatic Task Construction.}
After contact regularization, WorldComposer aligns the camera, robot base, and workspace into a shared coordinate frame using AprilTag~\cite{olson2011tags} and hand-eye calibration. This alignment ensures consistent camera views, robot coordinates, and object scales across simulation and the real world.
With the calibrated workspace in place, WorldComposer instantiates executable robot tasks by identifying and matching task-relevant objects in the real scene to high-fidelity simulation assets. We  categorize observed objects as rigid, articulated, or deformable. For rigid objects, we use SAM3D~\cite{sam3dteam2025sam3d3dfyimages} to reconstruct their geometry directly from real-world observations. For articulated and deformable objects, we match them to assets from our high-fidelity library using appearance correspondence (detailed in Appendix~\ref{app:asset_matching}). WorldComposer then samples initial object states and goal configurations, and defines success metrics and data-collection skills.

\subsection{Multi-Room Stitching}
\label{subsec:multi_room}

Since a 360-degree panorama typically covers only a single room, we need to stitch multiple room-level scenes into a house-scale, navigable environment. To this end, WorldComposer stitches independently generated scenes into a unified coordinate system for long-horizon tasks.

\textbf{Coarse Alignment via Panoramic Feature Matching.}
Stitching independently generated rooms is challenging because each room is reconstructed in its own local coordinate frame, and generated geometry often exhibits viewpoint-dependent artifacts away from the panorama center. WorldComposer estimates coarse room poses from feature matches between overlapping regions of the original panoramas. We extract local features using \textbf{SuperPoint}~\cite{detone2018superpoint} and match them with \textbf{LightGlue}~\cite{lindenberger2023lightglue}. The resulting correspondences are used to estimate an essential matrix, which is decomposed into a relative rotation $R_{ab}$ and a unit translation direction $\hat{t}_{ab}$. To resolve the scale ambiguity, we recover metric scale by matching the reconstructed camera-to-floor distance to the known camera height $h$:
\begin{equation}
t_{ab} = \alpha \hat{t}_{ab}, \quad \alpha = \frac{h}{\mathrm{median}(\{\mathbf{n}^\top p_i \mid p_i \in \mathcal{P}_{g}\})},
\end{equation}
where $\mathcal{P}_{g}$ denotes triangulated ground points and $\mathbf{n}$ is the estimated floor normal. This yields a metric-aware coarse transformation $T_{\mathrm{coarse}}=[R_{ab}|t_{ab}]$.

\textbf{Geometric Refinement via ICP.}
The coarse transformation provides a reliable initialization, but residual errors can still cause local discontinuities. WorldComposer refines the alignment by minimizing point-to-plane distances between dense overlapping point clouds $\mathcal{P}_a$ and $\mathcal{P}_b$ with ICP~\cite{CHEN1992ICP}:
\begin{equation}
E_{\mathrm{ICP}} = \sum_i \left\| (T_{\mathrm{fine}} \cdot p_a^i - p_b^j)^\top \mathbf{n}_b^j \right\|^2,
\end{equation}
where $T_{\mathrm{fine}}$ is initialized by $T_{\mathrm{coarse}}$. After convergence, the 3DGS kernels and collision meshes are merged into a unified coordinate system and exported as a stitched USD scene.

\textbf{Semantic Target Placement.}
For navigation tasks, WorldComposer uses Qwen2.5-VL~\cite{Qwen2.5-VL} to infer room types from rendered views of the stitched environment and propose poses for placing target objects in semantically appropriate regions of the reconstructed house.

\subsection{Generalizable Robot Learning and Evaluation}
\label{subsec:GRLE}

\textbf{Scalable Digital Cousins.}
The Digital Twin provides a task-ready reconstruction of a real-world scene but lacks the diversity needed for generalizable robot learning. To address this, WorldComposer generates Digital Cousins through scene and object variations while preserving task-relevant structure. 

For scenes, an LLM automatically generates editing prompts from a predefined template, guiding generative and reconstruction-based 3D world models to modify room style, texture, lighting, and semantic layout while preserving task-relevant structure. Generated scenes are further filtered for workspace validity and visual distinctness (detailed in Appendix~\ref{app:scene_cousin_generation}).

For objects, WorldComposer samples from our high-quality task-ready asset library, introducing assets of the same category with varied geometries, materials, and physical parameters.

\textbf{Automated Data Collection.}
For scalable data augmentation, WorldComposer provides an automated data collection mechanism based on parameterized scripted skills, enabling task trajectories to be generated from each simulated scene without repeated human teleoperation. The generated trajectories can be used alone or mixed with real-world demonstrations for Real2Sim2Real co-training.


Together, scalable Digital Cousins and automated data collection allow WorldComposer to serve as \textbf{a data engine for generalizable robot learning}, transforming finite real-world captures into diverse training trajectories that promote robust behaviors rather than overfitting to fixed configurations.  
         

Beyond data augmentation, our framework supports \textbf{a high-fidelity evaluation platform} that reflects real-world deployment. Rather than testing in a single static environment, we evaluate policies across hierarchical generalization tiers, from seen to unseen scene and object compositions. This provides a systematic test for zero-shot generalization that reliably correlates with real-world performance.

\section{Experiments}
\label{sec:experiments}

\subsection{Experimental Setup}
\label{subsec:setup}

\begin{wrapfigure}{r}{0.5\textwidth} 
  \centering
  \vspace{-5pt} 
  \includegraphics[width=\linewidth]{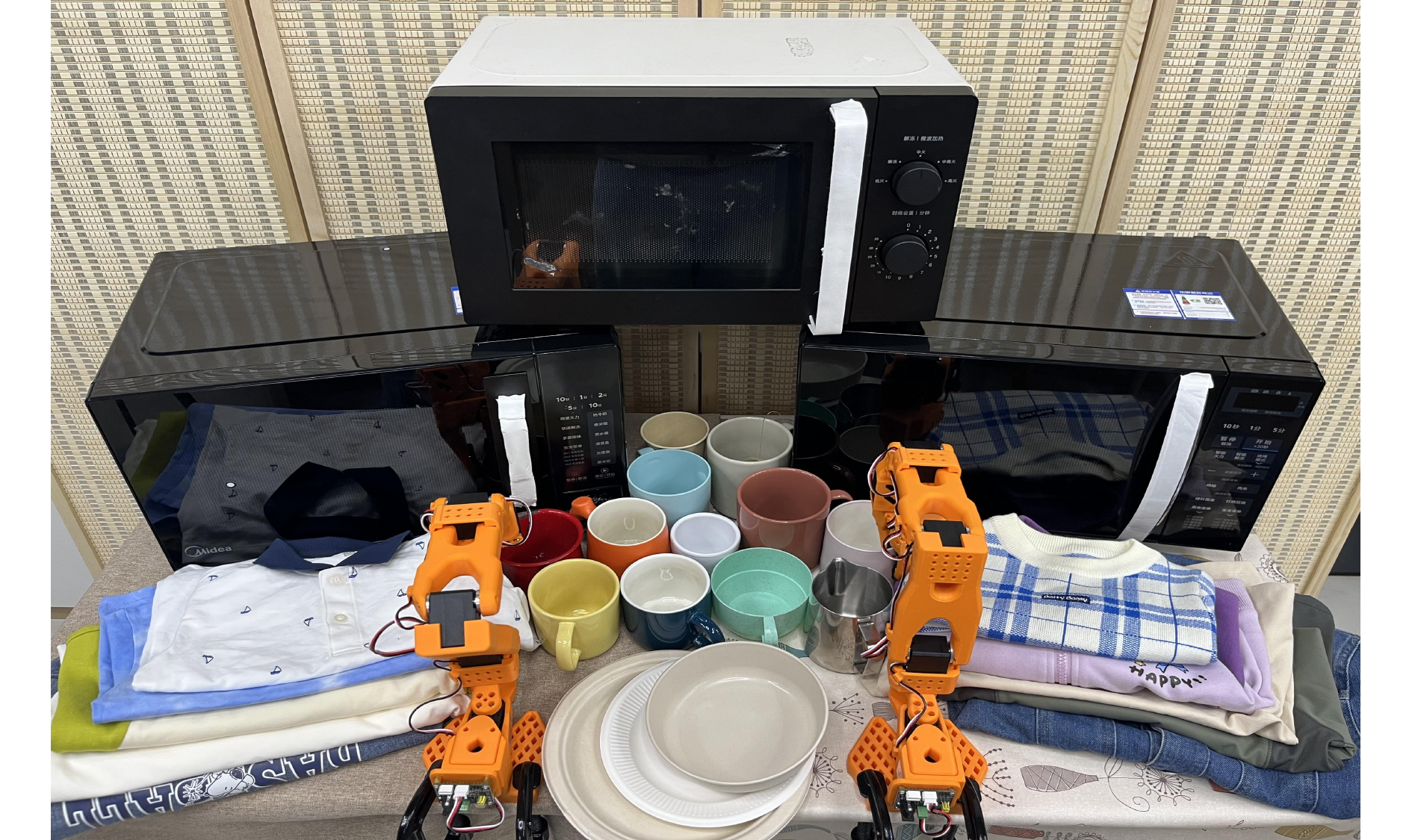}
  \caption{\textbf{Real-World Setup.} A bimanual LeRobot SO-101 platform with diverse task objects.}
  \label{fig:Realsetup}
  \vspace{-1em} 
\end{wrapfigure}

\textbf{Simulation and Hardware.}
We use \textbf{NVIDIA Isaac Sim} to instantiate all simulation environments and generate robot trajectories. The real-world setup consists of a bimanual system with two LeRobot SO-101 and diverse task objects, observed by a top-down RGB camera, as shown in Fig.~\ref{fig:Realsetup}.

\begin{figure*}[t]
\centering
\includegraphics[width=1.0\textwidth]{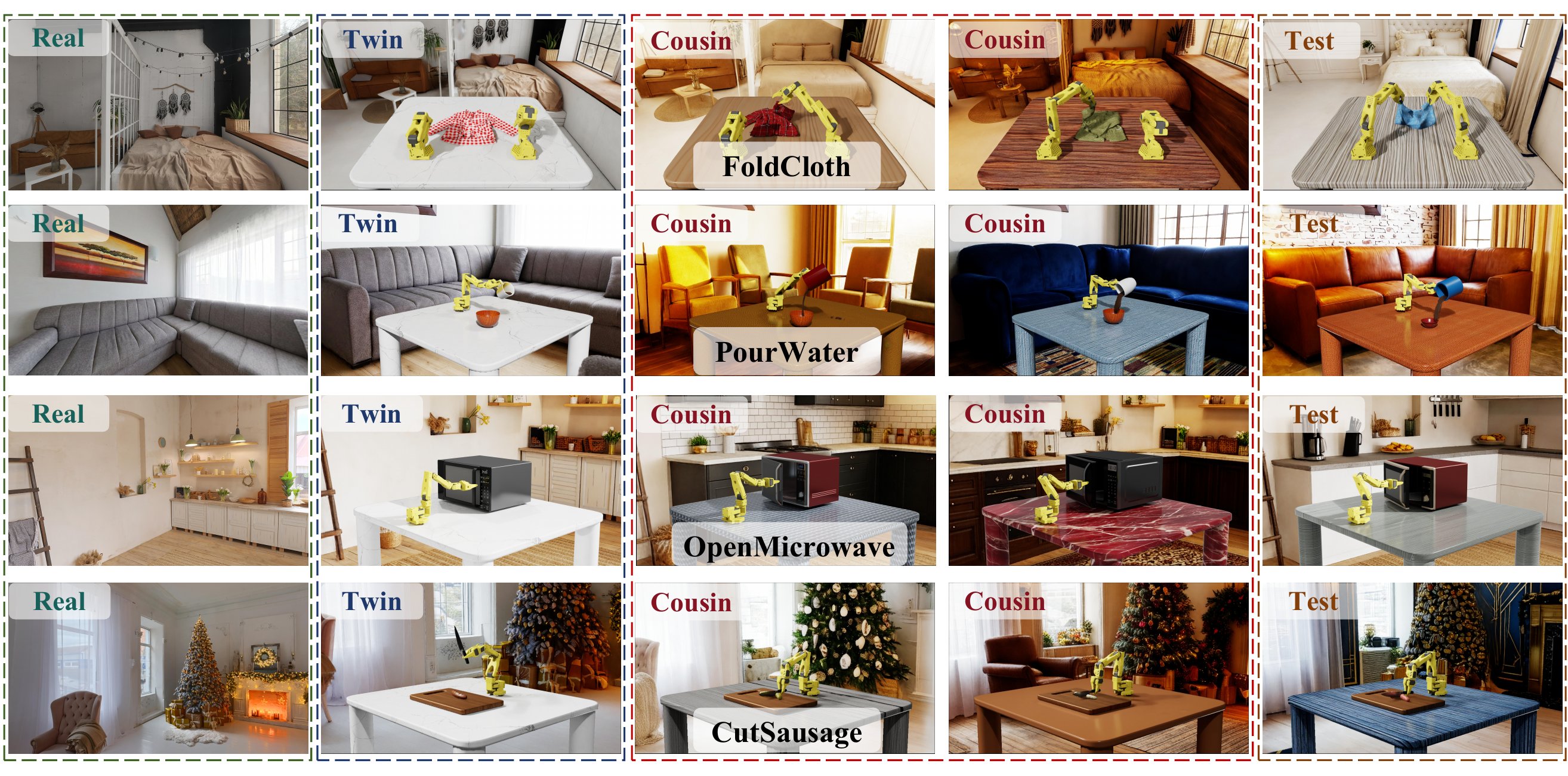}
\caption{\textbf{Simulation Tasks and Digital Cousin Generation.}
Our pipeline supports a diverse range of high-fidelity simulation tasks. Each row demonstrates the transition from a real-world capture (\textbf{Real}) to a digital reconstruction (\textbf{Twin}), followed by the generation of multiple \textbf{Digital Cousins}.}
\label{fig:SimTask}
\end{figure*}


\textbf{Tasks.}
As shown in Fig.~\ref{fig:SimTask}, our tasks span the main physical regimes supported by our asset library, including rigid-body placement, articulated-object operation, fluid interaction, and deformable manipulation. Detailed task definitions are provided in Appendix~\ref{app:task_details}.

\textbf{Evaluated Policies.}
The evaluated policies are ACT~\cite{zhao2023learning}, Diffusion Policy (DP)~\cite{chi2023diffusion}, SmolVLA~\cite{shukor2025smolvla}, and $\pi_0$~\cite{black2024pi_0}, spanning standard imitation-learning policies and VLA foundation models.
We assess data augmentation with DP and $\pi_0$, and sim-real correlation across all four policies.

\textbf{Baselines.}
We compare against two representative 3DGS-based Real2Sim2Real baselines, \textbf{GSWorld} and \textbf{DISCOVERSE}, as well as two augmentation baselines: \textbf{Domain Randomization (DR)} with randomized lighting and texture variations, and \textbf{Image Editing}. Because the baselines use different simulators, robots, and task protocols, we implement controlled adaptations under a unified evaluation setup: all baselines share our Isaac Sim tasks, LeRobot embodiment, and AutoCollection protocol. Thus, the comparison isolates how each method constructs task-ready simulation environments from real-world observations. Detailed reproduction procedures are provided in Appendix~\ref{app:baseline_details}.

\textbf{Data Collection \& Metrics.}
For standard baseline evaluations, we collect 50 expert demonstrations in both simulation and real-world settings.
For specific augmentation experiments, the exact training data compositions are detailed in their respective tables.
We conduct \textbf{100 trials} per configuration in simulation and \textbf{20 trials} in the real world.
We report the \textbf{Success Rate (SR)} across four generalization levels: \textit{Train}, \textit{Unseen Scene}, \textit{Unseen Object}, and \textit{Unseen Scene \& Object}.

\begin{figure*}[t]
\centering
\includegraphics[width=1.0\textwidth]{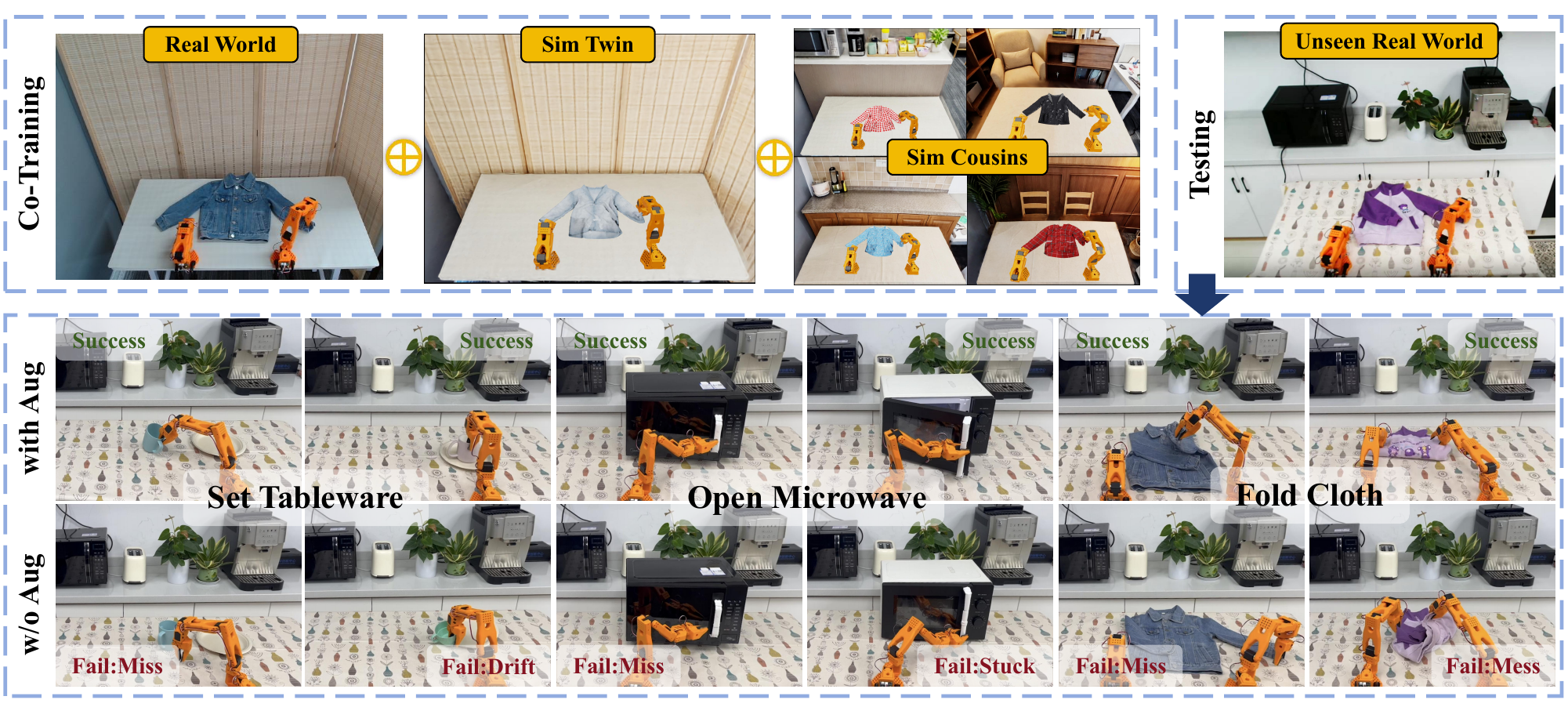}
\caption{\textbf{Real2Sim2Real Pipeline and Qualitative Analysis.}
(Top) The co-training framework combines real demonstrations with trajectories from Sim Twin and Cousins.
(Bottom) Comparison of policy execution on three tasks in unseen scenarios. Aug denotes the augmentation of generated data.}
\vspace{-2mm}
\label{fig:Real2sim2real}
\end{figure*}

\subsection{Efficacy of Digital Cousin Data Augmentation}
\label{subsec:augmentation}

We validate that our Digital Cousin data can effectively augment real-world demonstrations to improve policy generalization.
We conduct real-world experiments to answer: \textit{Does co-training with generated Digital Cousin data improve real-world policy robustness compared to existing data augmentation methods?}
Fig.~\ref{fig:Real2sim2real} shows our Real2Sim2Real co-training pipeline and qualitative comparisons.

\subsubsection{Real2Sim2Real Transfer}
\label{subsubsec:real2sim2real}

\textbf{Quality of Digital Twin Reconstruction.}
We first validate the fidelity of our reconstruction pipeline by comparing single-source baselines with 50 trajectories each.
As shown in Table~\ref{tab:real2sim2real_main} (top group), the policy trained on 50 Sim (Twin) trajectories achieves performance comparable to one trained on 50 Real trajectories (0.33 vs. 0.33 average success rate on unseen scenes and 0.25 vs. 0.27 on unseen objects).
This parity demonstrates that our Digital Twin reconstruction serves as a high-fidelity proxy for real-world interaction, establishing a solid foundation for data augmentation.

\textbf{Comparison with Data Augmentation Methods.}
We then evaluate various augmentation strategies under a fixed budget of 100 trajectories (Table~\ref{tab:real2sim2real_main}, bottom group).
Our method (50 Real + 50 Cousin) significantly outperforms all baselines under the same data budget.
Compared to 100 Real (0.37 scene / 0.35 object), our approach achieves average success rates of 0.55 / 0.48, representing a \textbf{49\% improvement in scene generalization} and a \textbf{37\% improvement in object generalization} while requiring only half the real-world data collection effort.

Notably, existing Real2Sim2Real frameworks GSWorld (0.38 / 0.33) and DISCOVERSE (0.40 / 0.35) provide only marginal gains over the 100 Real baseline. This is because these methods reconstruct a single Digital Twin without generating diverse variations lacking the Digital Cousin capability to systematically expand scene and object diversity. In contrast, our framework generates semantically coherent variations that expose the policy to a broader distribution of configurations.

\begin{table}[t!]
  \centering
  \caption{\textbf{Real-World Evaluation: Comparison with Baselines.}
  We compare Diffusion Policy trained with different data augmentation strategies.
  ``Twin'' denotes reconstructed simulation scenes, ``Cousin'' denotes generated variations, and ``DR'' denotes Domain Randomization.}
  \small
  \resizebox{\textwidth}{!}{
    \begin{tabular}{l *{8}{c}}
      \toprule
      & \multicolumn{2}{c}{Set Tableware} & \multicolumn{2}{c}{Open Microwave} & \multicolumn{2}{c}{Fold Cloth} & \multicolumn{2}{c}{Average} \\
      \cmidrule(lr){2-3} \cmidrule(lr){4-5} \cmidrule(lr){6-7} \cmidrule(lr){8-9}
      & Scene & Obj & Scene & Obj & Scene & Obj & Scene & Obj\\
      \midrule
      \multicolumn{9}{l}{\textit{Single-source baselines (50 trajectories)}} \\
      \quad 50 Real & 0.30 & 0.30 & 0.50 & 0.45 & 0.20 & 0.05 & 0.33 & 0.27 \\
      \quad 50 Sim Twin (Ours) & 0.40 & 0.30 & 0.45 & 0.35 & 0.15 & 0.10 & 0.33 & 0.25 \\
      \midrule
      \multicolumn{9}{l}{\textit{Augmentation methods (100 trajectories)}} \\
      \quad 100 Real & 0.30 & 0.40 & 0.50 & 0.45 & 0.30 & 0.20 & 0.37 & 0.35 \\
      \quad 50 Real + 50 Twin & 0.25 & 0.40 & 0.50 & 0.50 & 0.25 & 0.15 & 0.33 & 0.35 \\
      \quad 50 Real + 50 DR (Lighting) & 0.40 & 0.40 & 0.60 & 0.55 & \textbf{0.40} & 0.15 & 0.47 & 0.37 \\
      \quad 50 Real + 50 DR (Texture) & 0.30 & 0.45 & 0.65 & 0.45 & 0.30 & 0.20 & 0.42 & 0.37 \\
      \quad 50 Real + 50 DR (Lighting + Texture)
    & 0.45 & 0.45
    & 0.65 & 0.55
    & 0.35 & 0.20
    & 0.48 & 0.40 \\
    \quad 50 Real + 50 Sim (Image Editing)
    & 0.35 & 0.45
    & 0.60 & 0.50
    & 0.30 & 0.20
    & 0.42 & 0.38 \\
      \quad 50 Real + 50 Sim (GSWorld~\cite{jiang2025gsworldclosedloopphotorealisticsimulation}) & 0.40 & 0.35 & 0.45 & 0.50 & 0.30 & 0.15 & 0.38 & 0.33 \\
      \quad 50 Real + 50 Sim (DISCOVERSE~\cite{DISCOVERSE}) & 0.45 & 0.45 & 0.50 & 0.45 & 0.25 & 0.15 & 0.40 & 0.35 \\
      \quad 50 Real + 50 Sim Cousin (Scene) & 0.50 & 0.40 & \textbf{0.70} & 0.50 & \textbf{0.40} & 0.25 & 0.53 & 0.38 \\ 
      \quad 50 Real + 50 Sim Cousin (Object) & 0.30 & 0.45 & 0.55 & \textbf{0.60} & 0.35 & 0.30 & 0.40 & 0.45 \\
      \midrule
      \quad \textbf{50 Real + 50 Sim Cousin (Ours)} & \textbf{0.55} & \textbf{0.50} & \textbf{0.70} & \textbf{0.60} & \textbf{0.40} & \textbf{0.35} & \textbf{0.55} & \textbf{0.48} \\
      \bottomrule
    \end{tabular}
    \label{tab:real2sim2real_main}
  }
\end{table}

Our method also substantially surpasses DR-based augmentation (0.47 / 0.37 for lighting DR, 0.42 / 0.37 for texture DR, and 0.48 / 0.40 for combined lighting and texture DR) and image editing (0.42 / 0.38), demonstrating that structured scene and object generation provides richer visual and geometric variation for policy generalization.

Furthermore, the ablation on variation type reveals complementary benefits: Scene-only Cousins primarily improve scene generalization (0.53 / 0.38), Object-only Cousins primarily improve object generalization (0.40 / 0.45), and their combination achieves the best performance across both scene and object generalization settings (0.55 / 0.48).
The qualitative comparison in Fig.~\ref{fig:Real2sim2real} (bottom) further illustrates this: policies trained without augmentation fail in novel configurations due to scene overfitting, while our augmented policy successfully generalizes.

\begin{figure}[t!]
\centering
\includegraphics[width=1.0\textwidth]{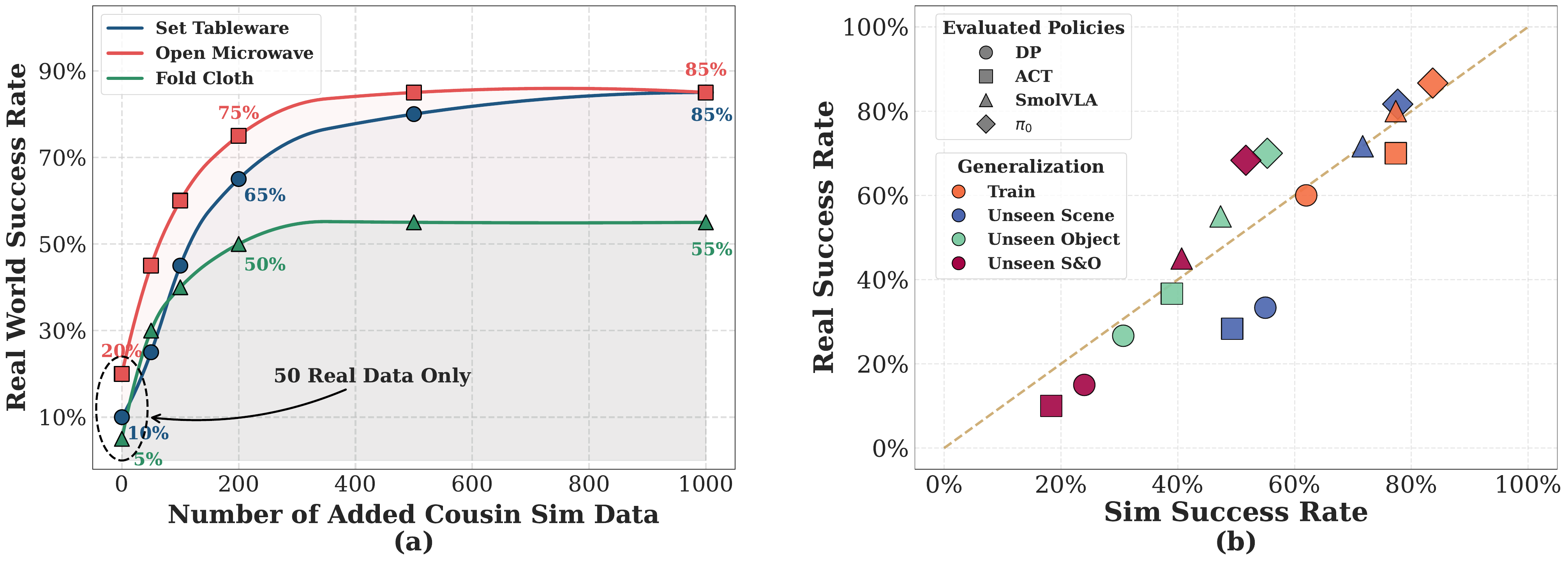}
\caption{
\textbf{(a) Performance with Increasing Cousin Sim Data.} Real-world success rates on Set Tableware, Open Microwave, and Fold Cloth with increasing Cousin Sim data under unseen scene and object variations.
\textbf{(b) Sim-Real Evaluation Correlation.} The scatter plot illustrates average success rates across three tasks for various policy methods under four generalization levels.
}

\label{fig:ScalingandCorrelationFig}
\vspace{-1em}
\end{figure}

\textbf{Scaling Analysis.}
We incrementally add up to 1,000 Digital Cousin trajectories to the 50 Real trajectories on the most challenging \textit{unseen scene \& object} setting.
As shown in Fig.~\ref{fig:ScalingandCorrelationFig}(a), we observe a consistent upward trend across all three tasks, demonstrating that large-scale diverse data generated by \textbf{WorldComposer} continues to boost policy generalization.

\subsubsection{Simulation Experiments}
\label{subsubsec:sim_validation}

To further validate the efficacy of Digital Cousin data, we conduct experiments purely in simulation across seven manipulation tasks.
As shown in Table~\ref{tab:dp_simulated_results} and Table~\ref{tab:vla_simulated_results}, adding Cousin data consistently boosts generalization for both Diffusion Policy (0.66 / 0.46 $\rightarrow$ 0.71 / 0.63, scene / object) and $\pi_0$ (0.78 / 0.55 $\rightarrow$ 0.83 / 0.76).
The improvements are particularly pronounced in physically complex tasks such as \textit{Pour Water} (object: 0.46 $\rightarrow$ 0.70), confirming that our generated data captures meaningful physical and visual diversity across different policy architectures.

\begin{table}[t!]
  \centering
  \caption{\textbf{Simulation Validation: Diffusion Policy.}
  We compare Diffusion Policy trained under two data settings on novel scenes and objects.}
  \small
  \begin{tabular*}{\textwidth}{@{\extracolsep{\fill}} l *{8}{c}}
    \toprule
    & \multicolumn{2}{c}{Set Tableware}
    & \multicolumn{2}{c}{Pour Water}
    & \multicolumn{2}{c}{Open Microwave}
    & \multicolumn{2}{c}{Close Drawer} \\
    \cmidrule(lr){2-3}  \cmidrule(lr){4-5}  \cmidrule(lr){6-7} \cmidrule(lr){8-9}
    & Scene & Object & Scene & Object & Scene & Object & Scene & Object\\
    \midrule
    100 Original data               & 0.61 & 0.30 & 0.64 & 0.46 & 0.59 & 0.33 & 0.78 & 0.48 \\
    100 Original + 200 Cousin data  & 0.68 & 0.50 & 0.70 & 0.70 & 0.62 & 0.56 & 0.87 & 0.72 \\
    \midrule
    & \multicolumn{2}{c}{Fold Cloth}
    & \multicolumn{2}{c}{Cut Sausage}
    & \multicolumn{2}{c}{Assemble Burger}
    & \multicolumn{2}{c}{Average} \\
    \cmidrule(lr){2-3}  \cmidrule(lr){4-5}  \cmidrule(lr){6-7} \cmidrule(lr){8-9}
    & Scene & Object & Scene & Object & Scene & Object & Scene & Object\\
    \midrule
    100 Original data               & 0.45 & 0.29 & 0.96 & 0.86 & 0.57 & 0.47 & 0.66 & 0.46 \\
    100 Original + 200 Cousin data  & 0.54 & 0.47 & 0.96 & 0.90 & 0.63 & 0.58 & 0.71 & 0.63 \\
    \bottomrule
  \end{tabular*}
  \label{tab:dp_simulated_results}
\end{table}

\begin{table}[t!]
  \centering
  \caption{\textbf{Simulation Validation: $\pi_0$ VLA.}
  We compare $\pi_0$ trained under two data settings on novel scenes and objects.}
  \small
  \begin{tabular*}{\textwidth}{@{\extracolsep{\fill}} l *{8}{c}}
    \toprule
    & \multicolumn{2}{c}{Set Tableware}
    & \multicolumn{2}{c}{Open Microwave}
    & \multicolumn{2}{c}{Fold Cloth}
    & \multicolumn{2}{c}{Average}\\
    \cmidrule(lr){2-3}  \cmidrule(lr){4-5}  \cmidrule(lr){6-7} \cmidrule(lr){8-9}
    & Scene & Object & Scene & Object & Scene & Object & Scene & Object \\
    \midrule
    100 Original data               & 0.84 & 0.53 & 0.82 & 0.58 & 0.67 & 0.55 & 0.78 & 0.55 \\
    100 Original + 200 Cousin data  & 0.89 & 0.78 & 0.88 & 0.81 & 0.72 & 0.69 & 0.83 & 0.76 \\
    \bottomrule
  \end{tabular*}
  \label{tab:vla_simulated_results}
  \vspace{-1em}
\end{table}

\subsection{Real-to-Sim Evaluation Correlation}
\label{subsec:correlation}

A critical prerequisite for a reliable simulation evaluation platform is its ability to precisely reflect real-world performance.
To validate the fidelity of our Digital Cousin environments, we conducted a rigorous correlation analysis by comparing success rates across three representative tasks.

As illustrated in Fig.~\ref{fig:ScalingandCorrelationFig}(b), we observe a strong positive linear relationship between simulation and real-world success rates.
The data points, representing ACT, DP, SmolVLA, and $\pi_0$ across multiple generalization levels, cluster tightly around the identity line (detailed results are presented in Appendix~\ref{app:D}).
The \textbf{Pearson correlation coefficient} of $r = 0.91$ demonstrates that our simulation evaluation aligns closely with real-world performance, with policy rankings and generalization trends consistently preserved across domains.
This establishes our system as a high-fidelity platform for evaluating policies at scale prior to real-world deployment.

\subsection{Navigation in Stitched Multi-Room Environments}
\label{subsec:nav_eval}

To evaluate scene-level interaction in stitched environments, we conduct \textbf{Zero-Shot Object-Goal Navigation (ZSON)} experiments with a movable XLeRobot. This task requires the robot to traverse complex, house-scale floorplans to locate interactive assets within a unified coordinate system.

\textbf{Setup.} Five distinct interactive assets from our library are procedurally injected into the stitched layout. We conduct 20 evaluation episodes per asset (100 total), with starting positions sampled from rooms distant to the goal to necessitate long-horizon, cross-room planning.

\textbf{Evaluated Method.} We use \textbf{VLFM}~\cite{yokoyama2024vlfm}, a zero-shot policy using a pretrained VLM to extract semantic cues from RGB streams, identifying promising frontiers in unseen environments.

\begin{figure*}[t] 
\centering

\begin{subfigure}[b]{0.48\textwidth}
\centering
\includegraphics[width=\linewidth]{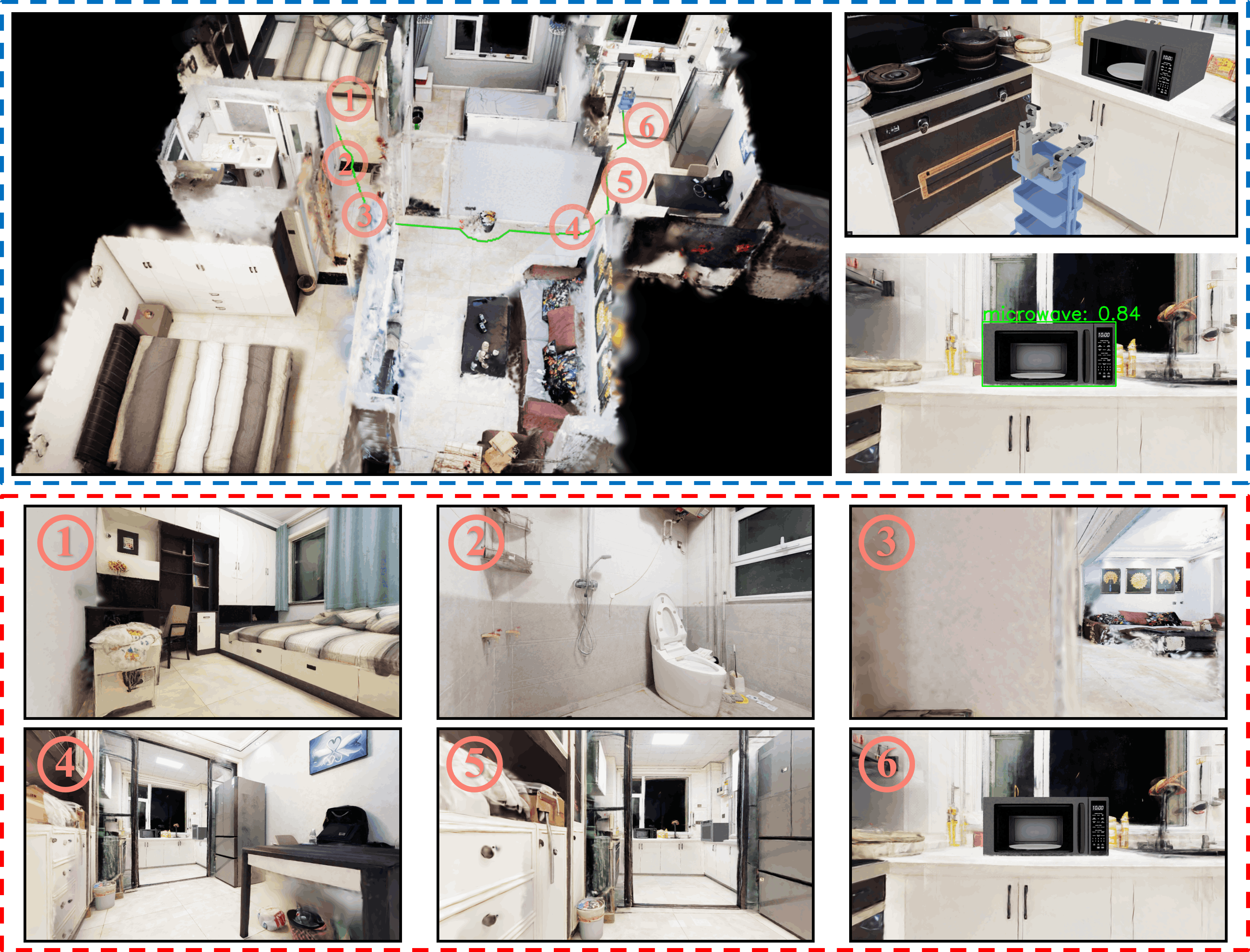}
\caption{ }
\label{fig:nav_sim}
\end{subfigure}
\hfill
\begin{subfigure}[b]{0.50\textwidth}
\centering
\includegraphics[width=\linewidth]{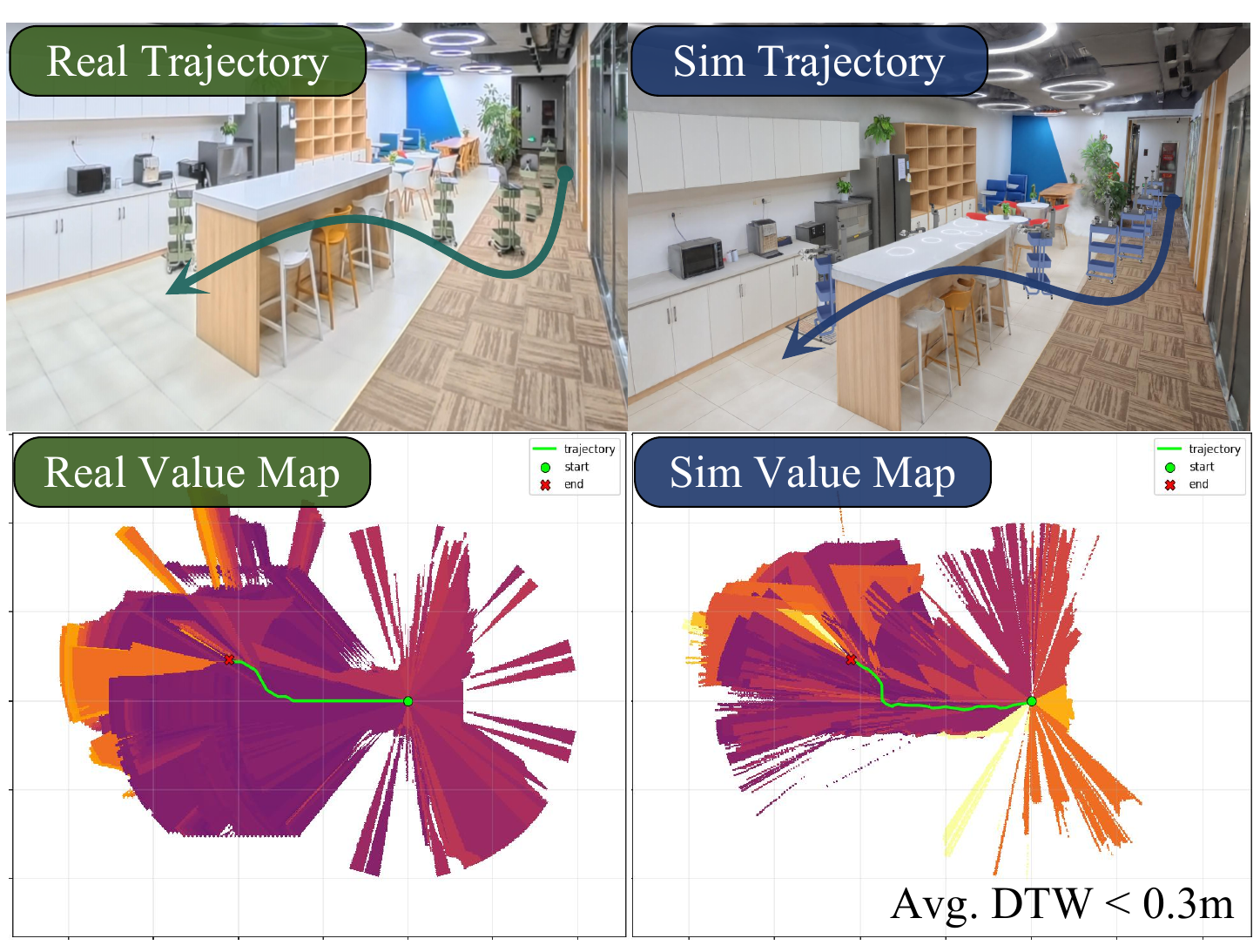}
\caption{ }
\label{fig:nav_real}
\end{subfigure}
\vspace{-2mm}
\caption{
\textbf{Multi-Room Navigation.}
\textbf{(a)} A zero-shot object-goal navigation episode in a stitched sim multi-room environment, with the trajectory and egocentric observations.
\textbf{(b)} Sim-real navigation comparison showing trajectories and value maps in a 1:1 real-world environment. Colors in the value maps indicate goal-relevance scores, ranging from dark purple (low) to pale yellow (high).}
\label{fig:combined_nav}
\vspace{-5mm} 
\end{figure*}

\begin{wraptable}{l}{0.39\textwidth}
\centering
\vspace{-4mm} 
\caption{\textbf{Navigation Results.} VLFM in stitched sim multi-room environment.}
\vspace{-1mm}
\label{tab:nav_results}
\small 
\begin{tabular}{lcc}
\toprule
Task (Target) & SR $\uparrow$ & SPL $\uparrow$ \\ \midrule
Microwave     & 0.70  & 0.58 \\
Chair         & 0.60  & 0.47 \\
Toilet        & 0.80  & 0.66 \\
Oven          & 0.55  & 0.42 \\
Refrigerator  & 0.75  & 0.61 \\ \midrule
\textbf{Total} & \textbf{0.68} & \textbf{0.55} \\ \bottomrule
\end{tabular}
\vspace{-5mm} 
\end{wraptable}

As detailed in Table~\ref{tab:nav_results}, VLFM achieves a 68\% average success rate, demonstrating that our stitched multi-room environments provide a robust and semantically consistent testbed for semantic reasoning and long-horizon navigation.

Fig.~\ref{fig:combined_nav}(a) visualizes a representative ZSON navigation episode in the stitched multi-room environment, including the cross-room trajectory (Top) and first-person observations during traversal (Bottom).  

To further assess sim-real consistency, we run the same VLFM policy in the real-world environment and its 1:1 simulation reconstructed by our WorldComposer. Fig.~\ref{fig:combined_nav}(b) presents the resulting navigation trajectories and value maps. The value maps exhibit similar spatial patterns, while the trajectories have an average Dynamic Time Warping (DTW) distance below 0.3 m, indicating consistent semantic and geometric navigation cues across domains.
\section{Conclusion} 
\label{sec:conclusion}

We presented an automated generative Real2Sim2Real framework that generates high-fidelity, interactive simulations with diverse digital cousins from real-world panoramas.
Improved policy robustness and strong real-sim correlation validate the platform for generalizable robot learning and evaluation.

\textbf{Limitations and Future Work.}
While the 3D world model generates the scene as a single mesh, we will further explore instance-level decomposition, leveraging 3D semantic segmentation to replace static elements. Additionally, to address texture discontinuities between independently generated rooms, we plan to investigate cross-scene radiance field fusion and end-to-end 3DGS optimization.



\clearpage
{
\small
\bibliographystyle{unsrt}
\bibliography{references}

\begin{thebibliography}{10}

\bibitem{intelligence2025pi05visionlanguageactionmodelopenworld}
Physical Intelligence, Kevin Black, Noah Brown, James Darpinian, Karan
  Dhabalia, Danny Driess, Adnan Esmail, Michael Equi, Chelsea Finn, Niccolo
  Fusai, Manuel~Y. Galliker, Dibya Ghosh, Lachy Groom, Karol Hausman, Brian
  Ichter, Szymon Jakubczak, Tim Jones, Liyiming Ke, Devin LeBlanc, Sergey
  Levine, Adrian Li-Bell, Mohith Mothukuri, Suraj Nair, Karl Pertsch, Allen~Z.
  Ren, Lucy~Xiaoyang Shi, Laura Smith, Jost~Tobias Springenberg, Kyle
  Stachowicz, James Tanner, Quan Vuong, Homer Walke, Anna Walling, Haohuan
  Wang, Lili Yu, and Ury Zhilinsky.
\newblock $\pi_{0.5}$: a vision-language-action model with open-world
  generalization, 2025.

\bibitem{bjorck2025gr00t}
Johan Bjorck, Fernando Casta{\~n}eda, Nikita Cherniadev, Xingye Da, Runyu Ding,
  Linxi Fan, Yu~Fang, Dieter Fox, Fengyuan Hu, Spencer Huang, et~al.
\newblock Gr00t n1: An open foundation model for generalist humanoid robots.
\newblock {\em arXiv preprint arXiv:2503.14734}, 2025.

\bibitem{cheang2025gr3technicalreport}
Chilam Cheang, Sijin Chen, Zhongren Cui, Yingdong Hu, Liqun Huang, Tao Kong,
  Hang Li, Yifeng Li, Yuxiao Liu, Xiao Ma, Hao Niu, Wenxuan Ou, Wanli Peng,
  Zeyu Ren, Haixin Shi, Jiawen Tian, Hongtao Wu, Xin Xiao, Yuyang Xiao, Jiafeng
  Xu, and Yichu Yang.
\newblock Gr-3 technical report, 2025.

\bibitem{NVIDIA_Isaac_Sim}
{NVIDIA}.
\newblock {Isaac Sim}.

\bibitem{genesis2026genesisworld}
Genesis~AI Team.
\newblock The role of simulation in scalable robotics, genesis world 1.0, and
  the path forward.
\newblock {\em Genesis AI Blog}, May 2026.

\bibitem{li2025lehome}
Zeyi Li, Jade Yang, Jingkai Xu, Shangbin Xie, Yuran Wang, Zhenhao Shen,
  Tianxing Chen, Yan Shen, Wenjun Li, Yukun Zheng, Chaorui Zhang, Ming Chen,
  Chen Xie, and Ruihai Wu.
\newblock Lehome: A simulation environment for deformable object manipulation
  in household scenarios.
\newblock In {\em IROS 2025 - 5th Workshop on RObotic MAnipulation of
  Deformable Objects: holistic approaches and challenges forward}, 2025.

\bibitem{chen2025robotwin}
Tianxing Chen, Zanxin Chen, Baijun Chen, Zijian Cai, Yibin Liu, Qiwei Liang,
  Zixuan Li, Xianliang Lin, Yiheng Ge, Zhenyu Gu, et~al.
\newblock Robotwin 2.0: A scalable data generator and benchmark with strong
  domain randomization for robust bimanual robotic manipulation.
\newblock {\em arXiv preprint arXiv:2506.18088}, 2025.

\bibitem{marble2025}
World Labs.
\newblock Marble, 2025.
\newblock Accessed: 2026-01-25.

\bibitem{hyworld22026}
Team HY-World.
\newblock Hy-world 2.0: A multi-modal world model for reconstructing,
  generating, and simulating 3d worlds.
\newblock {\em arXiv preprint}, 2026.

\bibitem{gigaworldteam2025gigaworld0worldmodelsdata}
GigaWorld Team, Angen Ye, Boyuan Wang, Chaojun Ni, Guan Huang, Guosheng Zhao,
  Haoyun Li, Jiagang Zhu, Kerui Li, Mengyuan Xu, Qiuping Deng, Siting Wang,
  Wenkang Qin, Xinze Chen, Xiaofeng Wang, Yankai Wang, Yu~Cao, Yifan Chang,
  Yuan Xu, Yun Ye, Yang Wang, Yukun Zhou, Zhengyuan Zhang, Zhehao Dong, and
  Zheng Zhu.
\newblock Gigaworld-0: World models as data engine to empower embodied ai,
  2025.

\bibitem{jiang2025gsworldclosedloopphotorealisticsimulation}
Guangqi Jiang, Haoran Chang, Ri-Zhao Qiu, Yutong Liang, Mazeyu Ji, Jiyue Zhu,
  Zhao Dong, Xueyan Zou, and Xiaolong Wang.
\newblock Gsworld: Closed-loop photo-realistic simulation suite for robotic
  manipulation, 2025.

\bibitem{wang2023robogen}
Yufei Wang, Zhou Xian, Feng Chen, Tsun-Hsuan Wang, Yian Wang, Katerina
  Fragkiadaki, Zackory Erickson, David Held, and Chuang Gan.
\newblock Robogen: Towards unleashing infinite data for automated robot
  learning via generative simulation, 2023.

\bibitem{Maddukuri2025SimandRealCA}
Abhiram Maddukuri, Zhenyu Jiang, Lawrence~Yunliang Chen, Soroush Nasiriany,
  Yuqi Xie, Yu~Fang, Wenqi Huang, Zu~Wang, Zhenjia Xu, Nikita Chernyadev, Scott
  Reed, Ken Goldberg, Ajay Mandlekar, Linxi~Jim Fan, and Yuke Zhu.
\newblock Sim-and-real co-training: A simple recipe for vision-based robotic
  manipulation.
\newblock {\em ArXiv}, abs/2503.24361, 2025.

\bibitem{chen2024urdformerpipelineconstructingarticulated}
Zoey Chen, Aaron Walsman, Marius Memmel, Kaichun Mo, Alex Fang, Karthikeya
  Vemuri, Alan Wu, Dieter Fox, and Abhishek Gupta.
\newblock Urdformer: A pipeline for constructing articulated simulation
  environments from real-world images, 2024.

\bibitem{qureshi2024splatsimzeroshotsim2realtransfer}
Mohammad~Nomaan Qureshi, Sparsh Garg, Francisco Yandun, David Held, George
  Kantor, and Abhisesh Silwal.
\newblock Splatsim: Zero-shot sim2real transfer of rgb manipulation policies
  using gaussian splatting, 2024.

\bibitem{DISCOVERSE}
Yufei Jia, Guangyu Wang, Yuhang Dong, Junzhe Wu, Yupei Zeng, Haonan Lin, Zifan
  Wang, Haizhou Ge, Weibin Gu, Kairui Ding, Zike Yan, Yunjie Cheng, Yue Li,
  Ziming Wang, Chuxuan Li, Wei Sui, Lu~Shi, Guanzhong Tian, Ruqi Huang, and
  Guyue Zhou.
\newblock Discoverse: Efficient robot simulation in complex high-fidelity
  environments, 2025.

\bibitem{jia2026gsplaygroundhighthroughputphotorealisticsimulator}
Yufei Jia, Heng Zhang, Ziheng Zhang, Junzhe Wu, Mingrui Yu, Zifan Wang, Dixuan
  Jiang, Zheng Li, Chenyu Cao, Zhuoyuan Yu, Xun Yang, Haizhou Ge, Yuchi Zhang,
  Jiayuan Zhang, Zhenbiao Huang, Tianle Liu, Shenyu Chen, Jiacheng Wang, Bin
  Xie, Xuran Yao, Xiwa Deng, Guangyu Wang, Jinzhi Zhang, Lei Hao, Zhixing Chen,
  Yuxiang Chen, Anqi Wang, Hongyun Tian, Yiyi Yan, Zhanxiang Cao, Yizhou Jiang,
  Hanyang Shao, Yue Li, Lu~Shi, Bokui Chen, Wei Sui, Hanqing Cui, Yusen Qin,
  Ruqi Huang, Lei Han, Tiancai Wang, and Guyue Zhou.
\newblock Gs-playground: A high-throughput photorealistic simulator for
  vision-informed robot learning, 2026.

\bibitem{kong2025hunyuanvideosystematicframeworklarge}
Weijie Kong, Qi~Tian, Zijian Zhang, Rox Min, Zuozhuo Dai, Jin Zhou, Jiangfeng
  Xiong, Xin Li, Bo~Wu, Jianwei Zhang, Kathrina Wu, Qin Lin, Junkun Yuan,
  Yanxin Long, Aladdin Wang, Andong Wang, Changlin Li, Duojun Huang, Fang Yang,
  Hao Tan, Hongmei Wang, Jacob Song, Jiawang Bai, Jianbing Wu, Jinbao Xue, Joey
  Wang, Kai Wang, Mengyang Liu, Pengyu Li, Shuai Li, Weiyan Wang, Wenqing Yu,
  Xinchi Deng, Yang Li, Yi~Chen, Yutao Cui, Yuanbo Peng, Zhentao Yu, Zhiyu He,
  Zhiyong Xu, Zixiang Zhou, Zunnan Xu, Yangyu Tao, Qinglin Lu, Songtao Liu, Dax
  Zhou, Hongfa Wang, Yong Yang, Di~Wang, Yuhong Liu, Jie Jiang, and Caesar
  Zhong.
\newblock Hunyuanvideo: A systematic framework for large video generative
  models, 2025.

\bibitem{openai2024sora}
OpenAI.
\newblock Sora, 2024.
\newblock Accessed: 2026-01-25.

\bibitem{openai2025sora2}
OpenAI.
\newblock Sora 2, 2025.
\newblock Accessed: 2026-01-25.

\bibitem{wan2025wanopenadvancedlargescale}
Team Wan, Ang Wang, Baole Ai, Bin Wen, Chaojie Mao, Chen-Wei Xie, Di~Chen,
  Feiwu Yu, Haiming Zhao, Jianxiao Yang, Jianyuan Zeng, Jiayu Wang, Jingfeng
  Zhang, Jingren Zhou, Jinkai Wang, Jixuan Chen, Kai Zhu, Kang Zhao, Keyu Yan,
  Lianghua Huang, Mengyang Feng, Ningyi Zhang, Pandeng Li, Pingyu Wu, Ruihang
  Chu, Ruili Feng, Shiwei Zhang, Siyang Sun, Tao Fang, Tianxing Wang, Tianyi
  Gui, Tingyu Weng, Tong Shen, Wei Lin, Wei Wang, Wei Wang, Wenmeng Zhou, Wente
  Wang, Wenting Shen, Wenyuan Yu, Xianzhong Shi, Xiaoming Huang, Xin Xu, Yan
  Kou, Yangyu Lv, Yifei Li, Yijing Liu, Yiming Wang, Yingya Zhang, Yitong
  Huang, Yong Li, You Wu, Yu~Liu, Yulin Pan, Yun Zheng, Yuntao Hong, Yupeng
  Shi, Yutong Feng, Zeyinzi Jiang, Zhen Han, Zhi-Fan Wu, and Ziyu Liu.
\newblock Wan: Open and advanced large-scale video generative models, 2025.

\bibitem{nvidia2025worldsimulationvideofoundation}
NVIDIA, :, Arslan Ali, Junjie Bai, Maciej Bala, Yogesh Balaji, Aaron Blakeman,
  Tiffany Cai, Jiaxin Cao, Tianshi Cao, Elizabeth Cha, Yu-Wei Chao, Prithvijit
  Chattopadhyay, Mike Chen, Yongxin Chen, Yu~Chen, Shuai Cheng, Yin Cui, Jenna
  Diamond, Yifan Ding, Jiaojiao Fan, Linxi Fan, Liang Feng, Francesco Ferroni,
  Sanja Fidler, Xiao Fu, Ruiyuan Gao, Yunhao Ge, Jinwei Gu, Aryaman Gupta,
  Siddharth Gururani, Imad~El Hanafi, Ali Hassani, Zekun Hao, Jacob Huffman,
  Joel Jang, Pooya Jannaty, Jan Kautz, Grace Lam, Xuan Li, Zhaoshuo Li,
  Maosheng Liao, Chen-Hsuan Lin, Tsung-Yi Lin, Yen-Chen Lin, Huan Ling, Ming-Yu
  Liu, Xian Liu, Yifan Lu, Alice Luo, Qianli Ma, Hanzi Mao, Kaichun Mo,
  Seungjun Nah, Yashraj Narang, Abhijeet Panaskar, Lindsey Pavao, Trung Pham,
  Morteza Ramezanali, Fitsum Reda, Scott Reed, Xuanchi Ren, Haonan Shao, Yue
  Shen, Stella Shi, Shuran Song, Bartosz Stefaniak, Shangkun Sun, Shitao Tang,
  Sameena Tasmeen, Lyne Tchapmi, Wei-Cheng Tseng, Jibin Varghese, Andrew~Z.
  Wang, Hao Wang, Haoxiang Wang, Heng Wang, Ting-Chun Wang, Fangyin Wei, Jiashu
  Xu, Dinghao Yang, Xiaodong Yang, Haotian Ye, Seonghyeon Ye, Xiaohui Zeng,
  Jing Zhang, Qinsheng Zhang, Kaiwen Zheng, Andrew Zhu, and Yuke Zhu.
\newblock World simulation with video foundation models for physical ai, 2025.

\bibitem{chen2025largevideoplannerenables}
Boyuan Chen, Tianyuan Zhang, Haoran Geng, Kiwhan Song, Caiyi Zhang, Peihao Li,
  William~T. Freeman, Jitendra Malik, Pieter Abbeel, Russ Tedrake, Vincent
  Sitzmann, and Yilun Du.
\newblock Large video planner enables generalizable robot control, 2025.

\bibitem{kim2025videofrom3d3dscenevideo}
Geonung Kim, Janghyeok Han, and Sunghyun Cho.
\newblock Videofrom3d: 3d scene video generation via complementary image and
  video diffusion models, 2025.

\bibitem{ren2025gen3c3dinformedworldconsistentvideo}
Xuanchi Ren, Tianchang Shen, Jiahui Huang, Huan Ling, Yifan Lu, Merlin
  Nimier-David, Thomas Müller, Alexander Keller, Sanja Fidler, and Jun Gao.
\newblock Gen3c: 3d-informed world-consistent video generation with precise
  camera control, 2025.

\bibitem{zhen2025tesseractlearning4dembodied}
Haoyu Zhen, Qiao Sun, Hongxin Zhang, Junyan Li, Siyuan Zhou, Yilun Du, and
  Chuang Gan.
\newblock Tesseract: Learning 4d embodied world models, 2025.

\bibitem{james2020rlbench}
Stephen James, Zicong Ma, David~Rovick Arrojo, and Andrew~J Davison.
\newblock Rlbench: The robot learning benchmark \& learning environment.
\newblock {\em IEEE Robotics and Automation Letters}, 5(2):3019--3026, 2020.

\bibitem{liu2023libero}
Bo~Liu, Yifeng Zhu, Chongkai Gao, Yihao Feng, Qiang Liu, Yuke Zhu, and Peter
  Stone.
\newblock Libero: Benchmarking knowledge transfer for lifelong robot learning.
\newblock {\em NeurIPS}, 2023.

\bibitem{szot2021habitat}
Andrew Szot, Alexander Clegg, Eric Undersander, Erik Wijmans, Yili Zhao, John
  Turner, Noah Maestre, Mustafa Mukadam, Devendra~Singh Chaplot, Oleksandr
  Maksymets, et~al.
\newblock Habitat 2.0: Training home assistants to rearrange their habitat.
\newblock {\em NeurIPS}, 2021.

\bibitem{li2021igibson}
Chengshu Li, Fei Xia, Roberto Mart{\'\i}n-Mart{\'\i}n, Michael Lingelbach,
  Sanjana Srivastava, Bokui Shen, Kent Vainio, Cem Gokmen, Gokul Dharan, Tanish
  Jain, et~al.
\newblock igibson 2.0: Object-centric simulation for robot learning of everyday
  household tasks.
\newblock {\em arXiv preprint arXiv:2108.03272}, 2021.

\bibitem{Mu_2025_CVPR}
Yao Mu, Tianxing Chen, Zanxin Chen, Shijia Peng, Zhiqian Lan, Zeyu Gao, Zhixuan
  Liang, Qiaojun Yu, Yude Zou, Mingkun Xu, Lunkai Lin, Zhiqiang Xie, Mingyu
  Ding, and Ping Luo.
\newblock Robotwin: Dual-arm robot benchmark with generative digital twins.
\newblock In {\em CVPR}, 2025.

\bibitem{nasiriany2024robocasa}
Soroush Nasiriany, Abhiram Maddukuri, Lance Zhang, Adeet Parikh, Aaron Lo,
  Abhishek Joshi, Ajay Mandlekar, and Yuke Zhu.
\newblock Robocasa: Large-scale simulation of everyday tasks for generalist
  robots.
\newblock {\em arXiv preprint arXiv:2406.02523}, 2024.

\bibitem{li2023behavior}
Chengshu Li, Ruohan Zhang, Josiah Wong, Cem Gokmen, Sanjana Srivastava, Roberto
  Mart{\'\i}n-Mart{\'\i}n, Chen Wang, Gabrael Levine, Michael Lingelbach,
  Jiankai Sun, et~al.
\newblock Behavior-1k: A benchmark for embodied ai with 1,000 everyday
  activities and realistic simulation.
\newblock In {\em Conference on Robot Learning}, pages 80--93. PMLR, 2023.

\bibitem{geng2025roboverse}
Haoran Geng, Feishi Wang, Songlin Wei, Yuyang Li, Bangjun Wang, Boshi An,
  Charlie~Tianyue Cheng, Haozhe Lou, Peihao Li, Yen-Jen Wang, et~al.
\newblock Roboverse: Towards a unified platform, dataset and benchmark for
  scalable and generalizable robot learning.
\newblock {\em arXiv preprint arXiv:2504.18904}, 2025.

\bibitem{guo2025genesismultimodaldrivingscene}
Xiangyu Guo, Zhanqian Wu, Kaixin Xiong, Ziyang Xu, Lijun Zhou, Gangwei Xu,
  Shaoqing Xu, Haiyang Sun, Bing Wang, Guang Chen, Hangjun Ye, Wenyu Liu, and
  Xinggang Wang.
\newblock Genesis: Multimodal driving scene generation with spatio-temporal and
  cross-modal consistency, 2025.

\bibitem{xian2023fluidlab}
Zhou Xian, Bo~Zhu, Zhenjia Xu, Hsiao-Yu Tung, Antonio Torralba, Katerina
  Fragkiadaki, and Chuang Gan.
\newblock Fluidlab: A differentiable environment for benchmarking complex fluid
  manipulation.
\newblock {\em arXiv preprint arXiv:2303.02346}, 2023.

\bibitem{wang2023thin}
Yian Wang, Juntian Zheng, Zhehuan Chen, Zhou Xian, Gu~Zhang, Chao Liu, and
  Chuang Gan.
\newblock Thin-shell object manipulations with differentiable physics
  simulations.
\newblock In {\em ICLR}, 2023.

\bibitem{seita2023learningrearrangedeformablecables}
Daniel Seita, Pete Florence, Jonathan Tompson, Erwin Coumans, Vikas Sindhwani,
  Ken Goldberg, and Andy Zeng.
\newblock {Learning to Rearrange Deformable Cables, Fabrics, and Bags with
  Goal-Conditioned Transporter Networks}.
\newblock In {\em ICRA}, 2021.

\bibitem{lu2024garmentlab}
Haoran Lu, Ruihai Wu, Yitong Li, Sijie Li, Ziyu Zhu, Chuanruo Ning, Yan Zhao,
  Longzan Luo, Yuanpei Chen, and Hao Dong.
\newblock Garmentlab: A unified simulation and benchmark for garment
  manipulation.
\newblock {\em NeurIPS}, 2024.

\bibitem{huang2021plasticinelab}
Zhiao Huang, Yuanming Hu, Tao Du, Siyuan Zhou, Hao Su, Joshua~B Tenenbaum, and
  Chuang Gan.
\newblock Plasticinelab: A soft-body manipulation benchmark with differentiable
  physics.
\newblock {\em arXiv preprint arXiv:2104.03311}, 2021.

\bibitem{heiden2021disect}
Eric Heiden, Miles Macklin, Yashraj Narang, Dieter Fox, Animesh Garg, and Fabio
  Ramos.
\newblock Disect: A differentiable simulation engine for autonomous robotic
  cutting.
\newblock {\em arXiv preprint arXiv:2105.12244}, 2021.

\bibitem{deitke2022phone2procbringingrobustrobots}
Matt Deitke, Rose Hendrix, Luca Weihs, Ali Farhadi, Kiana Ehsani, and Aniruddha
  Kembhavi.
\newblock Phone2proc: Bringing robust robots into our chaotic world, 2022.

\bibitem{kerbl20233dgaussiansplattingrealtime}
Bernhard Kerbl, Georgios Kopanas, Thomas Leimkühler, and George Drettakis.
\newblock 3d gaussian splatting for real-time radiance field rendering, 2023.

\bibitem{li2025robogsimreal2sim2realroboticgaussian}
Xinhai Li, Jialin Li, Ziheng Zhang, Rui Zhang, Fan Jia, Tiancai Wang, Haoqiang
  Fan, Kuo-Kun Tseng, and Ruiping Wang.
\newblock Robogsim: A real2sim2real robotic gaussian splatting simulator, 2025.

\bibitem{han2025re3simgeneratinghighfidelitysimulation}
Xiaoshen Han, Minghuan Liu, Yilun Chen, Junqiu Yu, Xiaoyang Lyu, Yang Tian,
  Bolun Wang, Weinan Zhang, and Jiangmiao Pang.
\newblock Re$^3$sim: Generating high-fidelity simulation data via
  3d-photorealistic real-to-sim for robotic manipulation, 2025.

\bibitem{jain2025polarisscalablerealtosimevaluations}
Arhan Jain, Mingtong Zhang, Kanav Arora, William Chen, Marcel Torne,
  Muhammad~Zubair Irshad, Sergey Zakharov, Yue Wang, Sergey Levine, Chelsea
  Finn, Wei-Chiu Ma, Dhruv Shah, Abhishek Gupta, and Karl Pertsch.
\newblock Polaris: Scalable real-to-sim evaluations for generalist robot
  policies, 2025.

\bibitem{abouchakra2025realissimbridgingsimtorealgap}
Jad Abou-Chakra, Lingfeng Sun, Krishan Rana, Brandon May, Karl Schmeckpeper,
  Niko Suenderhauf, Maria~Vittoria Minniti, and Laura Herlant.
\newblock Real-is-sim: Bridging the sim-to-real gap with a dynamic digital
  twin, 2025.

\bibitem{dan2025xsimcrossembodimentlearningrealtosimtoreal}
Prithwish Dan, Kushal Kedia, Angela Chao, Edward~Weiyi Duan, Maximus~Adrian
  Pace, Wei-Chiu Ma, and Sanjiban Choudhury.
\newblock X-sim: Cross-embodiment learning via real-to-sim-to-real, 2025.

\bibitem{pfaff2025scalablereal2simphysicsawareasset}
Nicholas Pfaff, Evelyn Fu, Jeremy Binagia, Phillip Isola, and Russ Tedrake.
\newblock Scalable real2sim: Physics-aware asset generation via robotic
  pick-and-place setups, 2025.

\bibitem{wu20253dgut}
Qi~Wu, Janick Martinez~Esturo, Ashkan Mirzaei, Nicolas Moenne-Loccoz, and Zan
  Gojcic.
\newblock 3dgut: Enabling distorted cameras and secondary rays in gaussian
  splatting.
\newblock {\em Conference on Computer Vision and Pattern Recognition (CVPR)},
  2025.

\bibitem{carion2025sam3segmentconcepts}
Nicolas Carion, Laura Gustafson, Yuan-Ting Hu, Shoubhik Debnath, Ronghang Hu,
  Didac Suris, Chaitanya Ryali, Kalyan~Vasudev Alwala, Haitham Khedr, Andrew
  Huang, Jie Lei, Tengyu Ma, Baishan Guo, Arpit Kalla, Markus Marks, Joseph
  Greer, Meng Wang, Peize Sun, Roman Rädle, Triantafyllos Afouras, Effrosyni
  Mavroudi, Katherine Xu, Tsung-Han Wu, Yu~Zhou, Liliane Momeni, Rishi Hazra,
  Shuangrui Ding, Sagar Vaze, Francois Porcher, Feng Li, Siyuan Li, Aishwarya
  Kamath, Ho~Kei Cheng, Piotr Dollár, Nikhila Ravi, Kate Saenko, Pengchuan
  Zhang, and Christoph Feichtenhofer.
\newblock Sam 3: Segment anything with concepts, 2025.

\bibitem{Ransac}
Martin~A. Fischler and Robert~C. Bolles.
\newblock Random sample consensus: {A} paradigm for model fitting with
  applications to image analysis and automated cartography.
\newblock {\em Commun. {ACM}}, 24(6):381--395, 1981.

\bibitem{olson2011tags}
Edwin Olson.
\newblock {AprilTag}: A robust and flexible visual fiducial system.
\newblock In {\em Proceedings of the {IEEE} International Conference on
  Robotics and Automation ({ICRA})}, pages 3400--3407. IEEE, May 2011.

\bibitem{sam3dteam2025sam3d3dfyimages}
SAM~3D Team, Xingyu Chen, Fu-Jen Chu, Pierre Gleize, Kevin~J Liang, Alexander
  Sax, Hao Tang, Weiyao Wang, Michelle Guo, Thibaut Hardin, Xiang Li, Aohan
  Lin, Jiawei Liu, Ziqi Ma, Anushka Sagar, Bowen Song, Xiaodong Wang, Jianing
  Yang, Bowen Zhang, Piotr Dollár, Georgia Gkioxari, Matt Feiszli, and
  Jitendra Malik.
\newblock Sam 3d: 3dfy anything in images.
\newblock 2025.

\bibitem{detone2018superpoint}
Daniel DeTone, Tomasz Malisiewicz, and Andrew Rabinovich.
\newblock Superpoint: Self-supervised interest point detection and description,
  2018.

\bibitem{lindenberger2023lightglue}
Philipp Lindenberger, Paul-Edouard Sarlin, and Marc Pollefeys.
\newblock {LightGlue: Local Feature Matching at Light Speed}.
\newblock In {\em ICCV}, 2023.

\bibitem{CHEN1992ICP}
Yang Chen and Gérard Medioni.
\newblock Object modelling by registration of multiple range images.
\newblock {\em Image and Vision Computing}, 10(3):145--155, 1992.
\newblock Range Image Understanding.

\bibitem{Qwen2.5-VL}
Shuai Bai, Keqin Chen, Xuejing Liu, Jialin Wang, Wenbin Ge, Sibo Song, Kai
  Dang, Peng Wang, Shijie Wang, Jun Tang, Humen Zhong, Yuanzhi Zhu, Mingkun
  Yang, Zhaohai Li, Jianqiang Wan, Pengfei Wang, Wei Ding, Zheren Fu, Yiheng
  Xu, Jiabo Ye, Xi~Zhang, Tianbao Xie, Zesen Cheng, Hang Zhang, Zhibo Yang,
  Haiyang Xu, and Junyang Lin.
\newblock Qwen2.5-vl technical report.
\newblock {\em arXiv preprint arXiv:2502.13923}, 2025.

\bibitem{zhao2023learning}
Tony~Z Zhao, Vikash Kumar, Sergey Levine, and Chelsea Finn.
\newblock Learning fine-grained bimanual manipulation with low-cost hardware.
\newblock {\em arXiv preprint arXiv:2304.13705}, 2023.

\bibitem{chi2023diffusion}
Cheng Chi, Zhenjia Xu, Siyuan Feng, Eric Cousineau, Yilun Du, Benjamin
  Burchfiel, Russ Tedrake, and Shuran Song.
\newblock Diffusion policy: Visuomotor policy learning via action diffusion.
\newblock {\em The International Journal of Robotics Research}, page
  02783649241273668, 2023.

\bibitem{shukor2025smolvla}
Mustafa Shukor, Dana Aubakirova, Francesco Capuano, Pepijn Kooijmans, Steven
  Palma, Adil Zouitine, Michel Aractingi, Caroline Pascal, Martino Russi,
  Andres Marafioti, et~al.
\newblock Smolvla: A vision-language-action model for affordable and efficient
  robotics.
\newblock {\em arXiv preprint arXiv:2506.01844}, 2025.

\bibitem{black2024pi_0}
Kevin Black, Noah Brown, Danny Driess, Adnan Esmail, Michael Equi, Chelsea
  Finn, Niccolo Fusai, Lachy Groom, Karol Hausman, Brian Ichter, et~al.
\newblock $pi\_0$: A vision-language-action flow model for general robot
  control.
\newblock {\em arXiv preprint arXiv:2410.24164}, 2024.

\bibitem{yokoyama2024vlfm}
Naoki Yokoyama, Sehoon Ha, Dhruv Batra, Jiuguang Wang, and Bernadette Bucher.
\newblock Vlfm: Vision-language frontier maps for zero-shot semantic
  navigation.
\newblock In {\em International Conference on Robotics and Automation (ICRA)},
  2024.

\bibitem{oquab2023dinov2}
Maxime Oquab, Timothée Darcet, Theo Moutakanni, Huy~V. Vo, Marc Szafraniec,
  Vasil Khalidov, Pierre Fernandez, Daniel Haziza, Francisco Massa, Alaaeldin
  El-Nouby, Russell Howes, Po-Yao Huang, Hu~Xu, Vasu Sharma, Shang-Wen Li,
  Wojciech Galuba, Mike Rabbat, Mido Assran, Nicolas Ballas, Gabriel Synnaeve,
  Ishan Misra, Herve Jegou, Julien Mairal, Patrick Labatut, Armand Joulin, and
  Piotr Bojanowski.
\newblock Dinov2: Learning robust visual features without supervision, 2023.

\bibitem{openai_chatgpt}
{OpenAI}.
\newblock Chatgpt.
\newblock \url{https://chatgpt.com}.

\bibitem{schoenberger2016sfm}
Johannes~Lutz Sch\"{o}nberger and Jan-Michael Frahm.
\newblock Structure-from-motion revisited.
\newblock In {\em Conference on Computer Vision and Pattern Recognition
  (CVPR)}, 2016.

\bibitem{kerbl3Dgaussians}
Bernhard Kerbl, Georgios Kopanas, Thomas Leimk{\"u}hler, and George Drettakis.
\newblock 3d gaussian splatting for real-time radiance field rendering.
\newblock {\em ACM Transactions on Graphics}, 42(4), July 2023.

\bibitem{xiang2024structured}
Jianfeng Xiang, Zelong Lv, Sicheng Xu, Yu~Deng, Ruicheng Wang, Bowen Zhang,
  Dong Chen, Xin Tong, and Jiaolong Yang.
\newblock Structured 3d latents for scalable and versatile 3d generation.
\newblock {\em arXiv preprint arXiv:2412.01506}, 2024.

\end{thebibliography}
}

\clearpage
\appendix

\section{More Technical Details}

\subsection{Real-to-Sim Construction}
\noindent \textbf{Camera-Robot-Workspace Calibration.}
For tabletop manipulation, we define the workspace frame using an AprilTag board placed on the support surface. Given calibrated camera intrinsics, detecting the board provides the camera-to-workspace transform. With reliable robot kinematics, we attach a marker to the end-effector and solve standard hand-eye calibration from multiple robot poses and camera observations, obtaining the camera-to-robot transform that, together with the camera-to-workspace transform, determines the robot-base-to-workspace transform. These procedures align the camera, robot base, object placement coordinates, and simulated workspace into a shared frame.

\noindent \textbf{Asset Matching for Articulated and Deformable Objects.}
\label{app:asset_matching}
For articulated and deformable objects, we perform retrieval within the same physical category rather than direct reconstruction, since generative reconstruction often does not produce simulator-ready joints, deformable topology, or physical parameters. We extract DINOv2~\cite{oquab2023dinov2} image embeddings and select the asset with the highest cosine similarity to the real object image. This retrieves the asset with the closest appearance-level correspondence, including shape, texture, and color. The selected asset is then scaled and aligned to the calibrated workspace before being instantiated in Isaac Sim.

\subsection{Digital Cousin Generation}
\label{app:scene_cousin_generation}

\noindent\textbf{Editing Prompt Generation.}
For scene variations, WorldComposer uses the following template
to generate editing prompts with an LLM, guiding the 3D world
model to modify the scene while preserving task-relevant structure.

\begingroup
\footnotesize
\begin{verbatim}
You are generating a scene-editing instruction for a 3D world model
that will produce a diverse variant of a robot-manipulation training scene.

CONTEXT
  Room type: {room_type}
  Original scene brief: {twin_scene_caption}
  Working surface: a large tabletop present in the scene.

EDITING DIMENSIONS (may combine one or more)
  1. layout    either (a) globally rearrange or replace background
               furniture and decor to change the overall spatial
               composition of the room, or (b) locally add / remove /
               replace a specific piece of furniture or decor
  2. style     change the overall room style / theme. Options include:
               modern minimalist, Scandinavian, industrial loft,
               rustic farmhouse, B&B / homestay warmth,
               Mediterranean, Art Deco, bohemian eclectic,
               traditional Chinese, Japanese zen
  3. texture   change wall paint, floor material, or non-tabletop
               furniture texture / material
  4. lighting  change lighting warmth, intensity, color, or direction.
               Options include: warm evening, cool morning daylight,
               dim ambient, bright overhead, golden hour,
               cinematic contrast

HARD CONSTRAINTS (violation -> invalid output)
  1. Do NOT change the tabletop's position, height, orientation, or dimensions.
  2. Produce a visually distinct scene. Avoid trivial or invisible changes.

OUTPUT
  A natural-language instruction of 1-3 sentences, in English,
  executable by a 3D world-model language editor.
\end{verbatim}
\endgroup

The generated instruction guides the 3D world model to produce updated Gaussian splats and a corresponding collision mesh, which is re-regularized following Sec.~\ref{subsec:scene_gen}.

\noindent\textbf{Filtering and Validation.}
WorldComposer filters generated scenes through two automatic
gates to check workspace usability and visual distinctness.

\noindent\textbf{Operation-region gate.}
We apply the Digital Twin construction pipeline to each generated
scene, using 12 rendered views for support-surface segmentation.
Scenes are discarded if segmentation scores fall below 0.5,
mean surface height fluctuation exceeds 0.5\,m,
RANSAC plane fitting fails, or local height variation remains
above 1\,cm after contact regularization.

\noindent\textbf{Visual-edit gate.}
We render the Digital Twin and generated scene from eight
identical camera poses and extract DINOv2 ViT-B/14 embeddings.
If the mean cosine similarity between paired embeddings exceeds
0.90, the scene is regenerated with a newly sampled editing
instruction.

\subsection{High-Fidelity Interactive Simulation}
\label{app:High-Fide}

To enable interactive simulation, WorldComposer integrates the reconstructed scenes with high-quality assets supported by high-fidelity physics solvers. In particular, we curate a task-ready asset library spanning three major interaction categories:

\begin{itemize}[leftmargin=*]
    \item \textbf{Rigid Objects:}
    Rigid assets are represented with clean visual meshes and decomposed collision geometry for stable contact simulation. For real-to-sim mapping, we use SAM3D to associate observed real objects with their simulation counterparts.
    \item \textbf{Articulated Objects:} 
    For objects with kinematic chains, we define joint limits and drive mechanisms. This enables realistic kinematic interactions, such as opening microwaves and closing drawers, while maintaining valid physical states throughout the interaction. 
    \item \textbf{Deformable Objects:} 
    Following the simulation engine in LeHome~\cite{li2025lehome}, we employ material-specific physics solvers to handle complex non-rigid dynamics. We use \textbf{Position-Based Dynamics} for garment-like objects, the \textbf{Finite Element Method} for elastic volumetric objects such as soft foods, and \textbf{Dynamic Grid Method} for fluids.
\end{itemize}

\noindent \textbf{Implementation of Fluid Interaction (Pour Water)} 
For complex physical interactions such as the Pour Water task, we simulate fluid-solid dynamics via Position-Based Dynamics (PBD) and render continuous liquid surfaces using Isosurfaces. The water volume is initialized by uniformly sampling particles inside the container's convex hull. A trial is considered successful if $> 60\%$ of the fluid particles enter the target receptacle's bounding box.



\section{More Experiment Implementation Details}

\subsection{Task Definitions}
\label{app:task_details}

Our tasks cover five representative physical interaction regimes:
\begin{itemize}[leftmargin=*]
    \item \textit{Rigid Body Manipulation:} The robot picks up diverse rigid objects, such as cups, mugs, or bowls, and places them onto a designated target plate.
    \item \textit{Articulated Object Interaction:} The robot interacts with articulated household objects, including opening a microwave by grasping its handle and closing an open drawer by pushing it.
    \item \textit{Fluid Interaction:} In \textit{Pour Water}, the robot pours liquid into a target cup while avoiding spillage.
    \item \textit{Hybrid Food Preparation:} The robot performs sequential food-preparation interactions, such as slicing a sausage and stacking a patty.
    \item \textit{Deformable Manipulation:} In \textit{Fold Cloth}, the robot folds a garment into the target configuration.
\end{itemize}

\subsection{Baseline Reproduction Details}
\label{app:baseline_details}

\noindent\textbf{\textit{Image Editing.}}
Image editing augments training observations through static
background replacement, without reconstructing 3D scene
geometry or supporting physical interaction.
We use ChatGPT~\cite{openai_chatgpt} to edit the original scene
image into five background variants per task, using the same
language prompts as our scene cousins.
For each training frame, we use SAM3 to segment the robot arm
and task objects and composite them onto an edited background.
Object variations and trajectories remain identical to those used in our WorldComposer setting.

\textbf{\textit{GSWorld.}}
GSWorld builds photorealistic Real2Sim environments using 3D Gaussian Splatting and physics-aware scene alignment. We reproduce its core components by constructing a metric-scale static 3DGS Digital Twin with COLMAP~\cite{schoenberger2016sfm} and scale calibration from tabletop fiducial tags, aligning the reconstructed scene, camera, workspace, and robot base to our Isaac Sim task frame, and manually introducing a planar operation surface as the physical workspace proxy for interaction.

\textbf{\textit{DISCOVERSE.}}
DISCOVERSE couples 3DGS scene rendering with mesh-based physical assets in a modular Real2Sim simulator. We adapt its asset-construction pipeline by first reconstructing the task scene with vanilla 3DGS~\cite{kerbl3Dgaussians} following its background-node design. For task-relevant rigid objects, such as tableware assets, we generate textured meshes with TRELLIS~\cite{xiang2024structured} and refine their appearance via Blender/PBR relighting under scene-matched illumination. The reconstructed background, object assets, operation plane, camera, and robot are then manually aligned to our task frame.

\subsection{Experiment Settings}
\noindent \textbf{Digital Cousin Data.} 
Digital Cousins feature two core augmentations: Scene Cousins (3D World Model-edited backgrounds) and Object Cousins (assets with varied geometries and physics), supplemented by lighting, pose, and texture randomizations. Each task contains 5 scene variants and 10 object variants. 

\noindent \textbf{Policy Training Configurations.} 
All policies are trained using the LeRobot framework with default optimizer, learning rate and so on. Training is conducted on NVIDIA RTX 4090 and A100 GPUs. The specific configurations for each policy are as follows:

\begin{itemize}
    \item ACT: Batch size 32, 50,000 training steps
    \item Diffusion Policy: Batch size 64, 50,000 training steps
    \item SmolVLA: Batch size 64, 50,000 training steps
    \item $\pi_0$(A100): Batch size 8, 30,000 training steps
\end{itemize}

All other hyperparameters follow the default LeRobot configurations for their respective policy types.

\subsection{Deployment and Computational Cost}
\label{app:computational_cost}

Deployment requires a panoramic camera for real scene capture.
Table~\ref{tab:computational_cost} summarizes the time and
computational resources required at each stage.
Capture and calibration are performed once per real setup,
while scene generation and task-ready post-processing are required
for each Digital Twin or scene cousin.
The resulting scenes can be reused across tasks and trajectory
collection. Object cousins are sampled from our task-ready asset
library without additional scene generation.

\begin{table}[htbp]
    \centering
    \caption{Approximate time and resource requirements
    for scene construction and automated trajectory collection.}
    \label{tab:computational_cost}
    \small
    \setlength{\tabcolsep}{3pt}
    \renewcommand{\arraystretch}{1.15}

    \begin{tabular}{
        @{}
        >{\raggedright\arraybackslash}
        m{\dimexpr0.43\linewidth-6\tabcolsep\relax}
        >{\raggedright\arraybackslash}m{0.18\linewidth}
        >{\raggedright\arraybackslash}m{0.19\linewidth}
        >{\raggedright\arraybackslash}m{0.20\linewidth}
        @{}
    }
        \toprule
        \textbf{Phase} &
        \textbf{Wall-clock time} &
        \textbf{Frequency} &
        \textbf{Resources} \\
        \midrule

        Real scene capture and calibration &
        $\approx$30 min &
        Once per real setup &
        Human \\

        Scene generation: Gaussian splats and collision mesh &
        $\approx$30 min / scene &
        Per twin or scene cousin &
        Cloud API \\

        Task-ready post-processing: segmentation, plane fitting,
        contact regularization, asset matching, and alignment &
        $\approx$10 min / scene &
        Per twin or scene cousin &
        Local GPU \\

        Automated trajectory collection &
        $\approx$0.6 min / trajectory &
        Per trajectory &
        $1\times$ RTX 4090; peak VRAM $\approx$6 GB \\

        \bottomrule
    \end{tabular}
    \vspace{-1em}
\end{table}

\section{Additional Task Visualization}

Due to space limitations in the main manuscript, we only visualized a subset of the simulation tasks. In this section, we present the visualizations for the remaining tasks supported by our pipeline: \textit{Set Tableware}, \textit{Close Drawer}, and \textit{Assemble Burger}.

\begin{figure}[!ht] 
\centering
\includegraphics[width=0.5\textwidth]{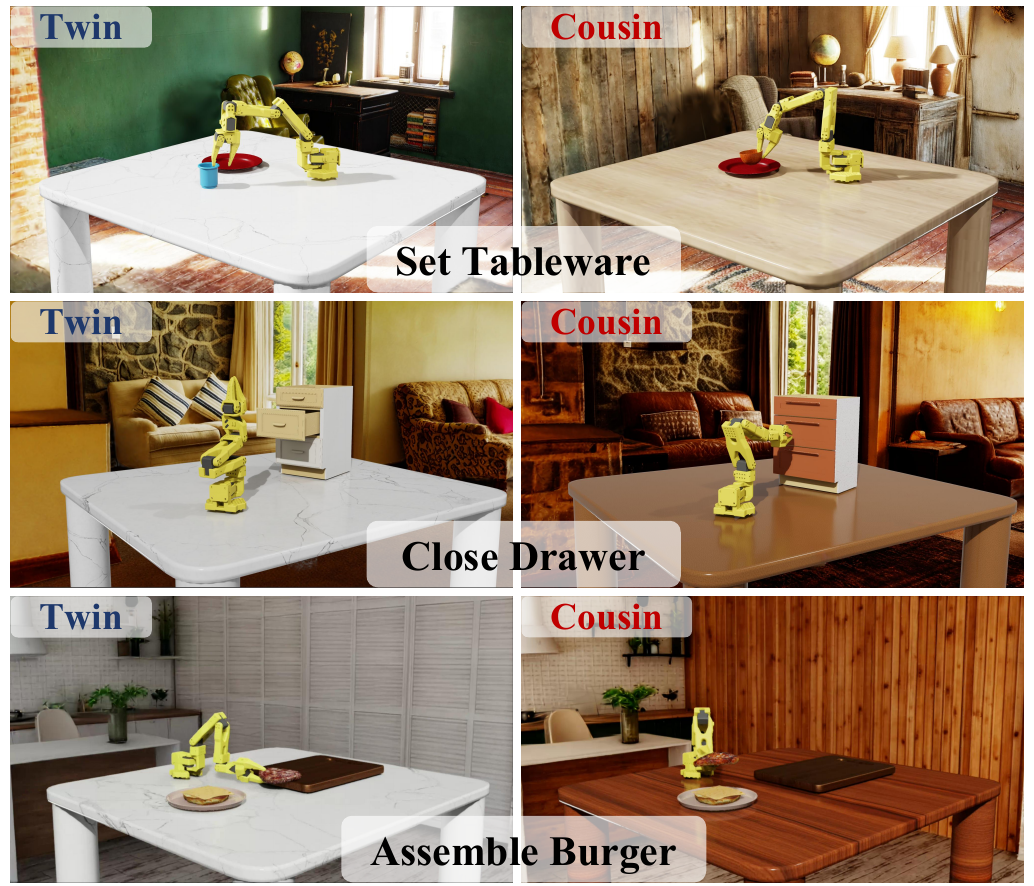} 
\caption{\textbf{Additional Simulation Tasks.} Visualization of the \textit{Digital Twin} (left) and generated \textit{Digital Cousin} (right) environments for the tasks not displayed in the main manuscript.}
\label{fig:supp_tasks}
\vspace{-1em}
\end{figure}

As shown in Fig.~\ref{fig:supp_tasks}, for each task, we display the \textbf{Digital Twin} (a precise reconstruction of the real-world setup) alongside a generated \textbf{Digital Cousin} (a semantically equivalent but visually diverse variation). These qualitative results further demonstrate the capabilities of our pipeline in generating high-fidelity and diverse simulation environments for a wide range of manipulation tasks.

\section{Detailed Quantitative Evaluation}
\label{app:D}

In this section, we provide the detailed quantitative results corresponding to the simulation-to-real correlation analysis presented in the main text. We report the detailed success rates for all baselines (ACT, DP, SmolVLA and $\boldsymbol{\pi_0}$) across three manipulation tasks: \textit{Set Tableware}, \textit{Open Microwave}, and \textit{Fold Cloth}.

\vspace{0.5em}
\noindent\textbf{Experimental Settings.} 
To ensure a rigorous and fair comparison, we standardized the data collection and evaluation protocols for both simulation and real-world experiments:

\begin{itemize}
    \item \textbf{Simulation:} We utilized a dataset comprising \textbf{100} expert trajectories per task for training. During the evaluation phase, each policy was tested over \textbf{100} independent trials for each generalization setting (Train, Unseen Scene, Unseen Object, and Unseen Scene \& Object) to report the average success rate.
    
    \item \textbf{Real-World:} We collected \textbf{50} high-quality human teleoperated trajectories for each task. For the real-robot evaluation, considering the operational costs, we conducted \textbf{20} evaluation trials for each method under each specific setting.
\end{itemize}

The specific breakdown of performance metrics is organized as follows: Table~\ref{tab:Evaluation1}, \ref{tab:Evaluation2}, \ref{tab:Evaluation3} present the simulation results, while Table~\ref{tab:Evaluation1_rw}, \ref{tab:Evaluation2_rw}, \ref{tab:Evaluation3_rw} detail the corresponding real-world performance.

\newcolumntype{C}{>{\centering\arraybackslash}X}
\begin{table}[htbp]
    \centering
    \caption{Simulation Evaluation (Set Tableware).}
    \small
    \begin{tabularx}{\linewidth}{@{}lCCCC@{}}
        \toprule
        Method \textbackslash{} Task Level & Train & Scene & Object & S \& O \\  
        \midrule
        ACT    & 0.83 & 0.54 & 0.36 & 0.13 \\
        DP  & 0.69 & 0.61 & 0.30 & 0.25 \\
        SmolVLA  & 0.85 & 0.78 & 0.49 & 0.40 \\
        $\boldsymbol{\pi_0}$ &  0.91 & 0.84 & 0.53 & 0.51 \\
        \bottomrule
    \end{tabularx}
    \vspace{-1em}
    \label{tab:Evaluation1}
\end{table}

\newcolumntype{C}{>{\centering\arraybackslash}X}
\begin{table}[htbp]
    \centering
    \caption{Simulation Evaluation (Open Microwave).}
    \small
    \begin{tabularx}{\linewidth}{@{}lCCCC@{}}
        \toprule
        Method \textbackslash{} Task Level & Train & Scene & Object & S \& O \\  
        \midrule
        ACT    & 0.91 & 0.61 & 0.46 & 0.23 \\
        DP  & 0.64 & 0.59 & 0.33 & 0.25 \\
        SmolVLA  & 0.82 & 0.76 & 0.49 & 0.44 \\
        $\boldsymbol{\pi_0}$  & 0.89 & 0.82 & 0.58 & 0.54 \\
        \bottomrule
    \end{tabularx}
    \vspace{-1em}
    \label{tab:Evaluation2}
\end{table}

\newcolumntype{C}{>{\centering\arraybackslash}X}
\begin{table}[htbp]
    \centering
    \caption{Simulation Evaluation (Fold Cloth).}
    \small
    \begin{tabularx}{\linewidth}{@{}lCCCC@{}}
        \toprule
        Method \textbackslash{} Task Level & Train & Scene & Object & S \& O \\  
        \midrule
        ACT    & 0.58 & 0.33 & 0.35 & 0.19 \\
        DP  & 0.53 & 0.45 & 0.29 & 0.22\\
        SmolVLA  & 0.65 & 0.61 & 0.44 & 0.38 \\
        $\boldsymbol{\pi_0}$  & 0.71 & 0.67 & 0.55 & 0.50 \\
        \bottomrule
    \end{tabularx}
    \vspace{-1em}
    \label{tab:Evaluation3}
\end{table}

\newcolumntype{C}{>{\centering\arraybackslash}X}
\begin{table}[!htbp]
    \centering
    \caption{Real-World Evaluation (Set Tableware).}
    \small
    \begin{tabularx}{\linewidth}{@{}lCCCC@{}}
        \toprule
        Method \textbackslash{} Task Level & Train & Scene & Object & S \& O \\  
        \midrule
        ACT  & 0.70  & 0.25 & 0.40 & 0.05 \\
        DP  &  0.60 & 0.30 & 0.30 & 0.10  \\
        SmolVLA  & 0.95 & 0.85 & 0.70 & 0.55  \\
        $\boldsymbol{\pi_0}$  & 0.95 & 0.90 & 0.90 & 0.85 \\
        \bottomrule
    \end{tabularx}
    \label{tab:Evaluation1_rw}
\end{table}

\newcolumntype{C}{>{\centering\arraybackslash}X}
\begin{table}[!htbp]
    \centering
    \caption{Real-World Evaluation (Open Microwave).}
    \small
    \begin{tabularx}{\linewidth}{@{}lCCCC@{}}
        \toprule
        Method \textbackslash{} Task Level & Train & Scene & Object & S \& O \\  
        \midrule
        ACT  & 0.90  & 0.45 & 0.60 & 0.25  \\
        DP & 0.75 & 0.50 & 0.45 & 0.35 \\
        SmolVLA  & 1.00 & 0.95 & 0.80 & 0.75 \\
        $\boldsymbol{\pi_0}$  & 1.00 & 0.95 & 0.85 & 0.85 \\
        \bottomrule
    \end{tabularx}
    \label{tab:Evaluation2_rw}
\end{table}

\newcolumntype{C}{>{\centering\arraybackslash}X}
\begin{table}[!htbp]
    \centering
    \caption{Real-World Evaluation (Fold Cloth).}
    \small
    \begin{tabularx}{\linewidth}{@{}lCCCC@{}}
        \toprule
        Method \textbackslash{} Task Level & Train & Scene & Object & S \& O \\  
        \midrule
        ACT  & 0.50  & 0.15 & 0.10 & 0.00  \\
        DP  &  0.45 & 0.20 & 0.05 & 0.00  \\
        SmolVLA  & 0.45 & 0.35 & 0.15 &  0.05 \\
        $\boldsymbol{\pi_0}$  & 0.65 & 0.60 & 0.35 & 0.35 \\
        \bottomrule
    \end{tabularx}
    \label{tab:Evaluation3_rw}
\end{table}

\section{More Experiments}

\subsection{Effect of Digital Cousin Diversity}
\label{app:cousin_diversity}

With the training budget fixed at 50 real and 50 cousin trajectories, we vary the number $N$ of scene or object variants. Scene and object cousins are evaluated on unseen scenes and unseen objects, respectively. As shown in Table~\ref{tab:cousin_diversity}, increasing $N$ from 1 to 10 raises average unseen-scene success from 0.40 to 0.62 and unseen-object success from 0.38 to 0.45.

\begin{table}[H]
    \centering
    \caption{\textbf{Effect of Digital Cousin Diversity.} Real-world success rates across three tasks under a fixed 100-trajectory training budget.}
    \label{tab:cousin_diversity}
    \small
    \setlength{\tabcolsep}{4pt}
    \renewcommand{\arraystretch}{1.1}
    \begin{tabular*}{\linewidth}{@{\extracolsep{\fill}}lccccc@{}}
        \toprule
        \textbf{Cousin} & \textbf{$N$} & \textbf{Set Tableware} & \textbf{Open Microwave} & \textbf{Fold Cloth} & \textbf{Average} \\
        \midrule
        Scene & 1  & 0.35 & 0.55 & 0.30 & 0.40 \\
        Scene & 5  & 0.50 & 0.70 & 0.40 & 0.53 \\
        Scene & 10 & 0.55 & 0.75 & 0.55 & 0.62 \\
        \midrule
        Object & 1  & 0.40 & 0.55 & 0.20 & 0.38 \\
        Object & 5  & 0.45 & 0.55 & 0.25 & 0.42 \\
        Object & 10 & 0.45 & 0.60 & 0.30 & 0.45 \\
        \bottomrule
    \end{tabular*}
\end{table}

\subsection{Real-World Navigation Alignment}
\label{app:real_nav_alignment}

To further validate the stitched multi-room environments, we conduct a 1:1 sim-real navigation comparison. We sample portal panoramas as visual anchors for robust stitching, enabling overlapping regions to compensate for reconstruction artifacts. As shown in Table~\ref{tab:RealNavi_appendix}, the simulated and real-world environments achieve comparable zero-shot navigation success rates. 

\begin{table}[H]
\centering
\caption{\textbf{Sim-Real Navigation.} Zero-shot success rates in the 1:1 multi-room environment.}
\label{tab:RealNavi_appendix}
\renewcommand{\arraystretch}{1.2}
\begin{tabular}{lc}
\toprule
\textbf{Scene} & \textbf{Success Rate (SR)} \\
\midrule
Sim & 12/20 \\
Real & 10/20 \\
\bottomrule
\end{tabular}
\end{table}

\end{document}